\documentclass[runningheads]{llncs}

\usepackage{pifont}
\usepackage{wrapfig}
\usepackage{xcolor}
\usepackage{lipsum}
\usepackage{xspace}
\usepackage{graphicx,float,subfig}
\usepackage{amsmath,amssymb,amsfonts,dsfont,pifont,bm,bbm,mathrsfs,mathtools,nicefrac}
\usepackage{algorithm,algpseudocode,listings}
\usepackage{booktabs,multirow,adjustbox,diagbox,threeparttable}
\usepackage[misc]{ifsym}
\definecolor{citeblue}{RGB}{48,111,186}
\usepackage[pagebackref=false,breaklinks=true,colorlinks=true,citecolor=citeblue,bookmarks=false]{hyperref}
\usepackage{cleveref}  
\usepackage[hypcap=false]{caption}
\crefname{section}{Sec.}{Secs.}
\Crefname{section}{Section}{Sections}
\crefname{table}{Tab.}{Tabs.}
\Crefname{table}{Table}{Tables}
\crefname{figure}{Fig.}{Figs.}
\Crefname{figure}{Figure}{Figures}
\crefname{equation}{Eq.}{Eqs.}
\Crefname{equation}{Equation}{Equations}
\newcommand\nonumfootnote[1]{%
\begingroup%
    \renewcommand\thefootnote{}\footnote{\hspace{-4pt}#1}%
    \addtocounter{footnote}{-1}%
\endgroup%
}

\begin{document}

\title{ADTR: Anomaly Detection Transformer with Feature Reconstruction}
\titlerunning{ADTR: Anomaly Detection Transformer with Feature Reconstruction}
\author{\hspace{-5pt} Zhiyuan You\inst{1} \and Kai Yang\inst{2} \and Wenhan Luo\inst{3} \and Lei Cui\inst{4} \and Yu Zheng\inst{1} \and Xinyi Le\inst{1}\Letter}
\authorrunning{You et al.}
\institute{
$^1$Shanghai Jiao Tong University,\quad $^2$SenseTime Research \\ $^3$Sun Yat-sen University,\quad $^4$Tsinghua University
}

\maketitle
\centerline{
\email{zhiyuanyou@foxmail.com}, \email{lexinyi@sjtu.edu.cn}
}

\begin{abstract}

Anomaly detection with only prior knowledge from normal samples attracts more attention because of the lack of anomaly samples. Existing CNN-based pixel reconstruction approaches suffer from two concerns. First, the reconstruction source and target are raw pixel values that contain indistinguishable semantic information. Second, CNN tends to reconstruct both normal samples and anomalies well, making them still hard to distinguish. In this paper, we propose Anomaly Detection TRansformer (ADTR) to apply a transformer to reconstruct pre-trained features. The pre-trained features contain distinguishable semantic information. Also, the adoption of transformer limits to reconstruct anomalies well such that anomalies could be detected easily once the reconstruction fails. Moreover, we propose novel loss functions to make our approach compatible with the normal-sample-only case and the anomaly-available case with both image-level and pixel-level labeled anomalies. The performance could be further improved by adding simple synthetic or external irrelevant anomalies. 
Extensive experiments are conducted on anomaly detection datasets including MVTec-AD and CIFAR-10. Our method achieves superior performance compared with all baselines. 


\nonumfootnote{\scriptsize{This work is sponsored by the Shanghai Foundation for Development of Science and Technology (21SQBS01502) and National Natural Science Foundation of China (62176152).}}

\end{abstract}

\section{Introduction}\label{sec:intro}


Unsupervised anomaly detection~\cite{bergmann2019mvtec,borghesi2019anomaly,gt} aims to identify anomalies using prior knowledge from only normal samples. Due to the extreme lack of anomalies in production lines, anomaly detection is attracting more and more interests. 

From the view of statistics, anomalies may be seen as distribution outliers of normal samples. In this setting, CNN-based reconstruction models like Auto-Encoder (AE), Variational Auto-Encoder (VAE), and Generative Adversarial Network (GAN) are usually adopted to model the distribution of normal samples~\cite{borghesi2019anomaly,dehaene2020iterative,gong2019memorizing,gradcon}. These methods train a model with only normal samples based on the assumption of generalization gap, which means that the reconstruction succeeds with only normal samples but fails with anomalies. The anomaly detection is performed with a distance metric between a sample and its reconstruction.

\begin{figure}[tb]
    \centering
    \includegraphics[width=1\linewidth]{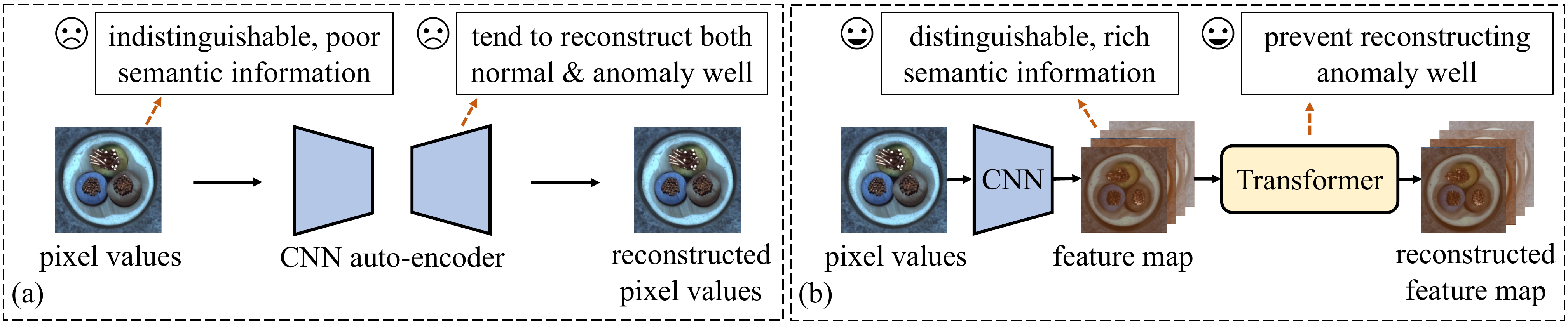}
    \vspace{-18pt}
    \caption{\textbf{(a) CNN-based pixel reconstruction methods} tend to reconstruct both normal samples and anomalies well, making them still hard to distinguish. Also, the pixel values contain indistinguishable semantic information. \textbf{(b) Our method} reconstructs features with distinguishable semantic information. Besides, the adoption of transformer limits the reconstruction of anomalies.}
\label{fig:abstract}
\vspace{-16pt}
\end{figure}

As show in \cref{fig:abstract}a, one concern about these approaches is the poor representation ability. The reconstruction targets are raw pixel values with poor semantic information. Therefore, these pixel reconstruction approaches usually fails when normal and anomalous regions share similar pixel values but different semantic information like different textures. In another aspect, it has been verified that the feature extractor pre-trained on large public datasets could extract distinguishable features for normal samples and anomalies~\cite{ts,kdad}. Thus we propose to reconstruct pre-trained features instead of raw pixel values. 

Taking CNN as the reconstruction model brings another issue (\cref{fig:abstract}a). CNN tends to take shortcuts to learn a somewhat ``identical mapping'', which means the anomalous regions are also reconstructed quite well~\cite{gong2019memorizing}. The great success of transformer in computer vision inspires us to propose a transformer-based reconstruction model. The query embedding in attention layer of transformer could limit the tendency of ``identical mapping'', which helps distinguish normal samples and anomalies (See \cref{subsec:prevent}). 

Besides, more anomaly samples are available with the runs of production lines~\cite{ts}, bringing anomaly detection the demands of compatibility with both the normal-sample-only case (only normal samples are available) and the anomaly-available case (normal samples and a few anomalies are available). 
Therefore, a unified approach that is compatible with both cases would be a better solution. 

In this paper, we propose a concise but powerful transformer-based anomaly detection approach. As shown in \cref{fig:abstract}b, a frozen pre-trained CNN backbone is adopted to extract features, then a transformer is used for feature reconstruction. The proposed approach has strong representation abilities, and could limit the tendency of ``identical mapping''. Moreover, novel loss functions are proposed for the compatibility with the anomaly-available case. The performance could be further improved by adding simple synthetic or external irrelevant anomalies. Our approach achieves state-of-the-art anomaly detection performance in anomaly detection datasets including MVTec-AD~\cite{bergmann2019mvtec} and CIFAR-10~\cite{krizhevsky2009learning}. 


\section{Related Work}\label{sec:related}


Existing anomaly detection approaches could be generally divided into two categories: reconstruction-based ones and projection-based ones. 

\textbf{Reconstruction-based approaches} assume that the reconstruction model trained with normal samples has a generalization gap with anomalies, thus fails to reconstruct anomalies. 
AE~\cite{ssimae,dehaene2020iterative,gong2019memorizing,park2020learning} and GAN~\cite{ocgan,sabokrou2018adversarially,zaheer2020old} are intuitive choices of reconstruction models. Zhou et al.~\cite{pnet} and Xia et al.~\cite{xia2020synthesize} respectively adopt the structural information and semantic segmentation information for better reconstruction. Zaheer et al.~\cite{zaheer2020old} utilize a discriminator to distinguish good or bad quality of reconstruction, and the predicted possibility of bad quality serves as an anomaly score. Gong et al.~\cite{gong2019memorizing} and Park et al.~\cite{park2020learning} introduce a memory module to select the most similar embedding in embedding storage of normal samples to restrict the generalization on anomalies. Dehaene et al.~\cite{vae-grad} refine the selection method with an iterative gradient-based approach. 

\textbf{Projection-based approaches} project samples into an embedding space, where normal samples and anomalies are more distinguishable. SVDD~\cite{svdd} extracts feature representation with the one-class classification objective. Yi and Yoon~\cite{yi2020patch} propose a patch-based SVDD with multiple kernels. Liu et al.~\cite{vevae} and Kwon et al.~\cite{gradcon} find that the back-propagated gradients of normal samples and anomalies are more distinguishable. FCDD~\cite{liznerski2020explainable} is trained to enlarge the embedding differences between normal samples and anomalies, where the mapped samples are themselves an explanation heat map. Bergmann et al.~\cite{ts} utilize a teacher-student network, assuming that the embedding differences between normal samples and anomalies would be enlarged through knowledge distillation. Salehi et al.~\cite{kdad} extend the knowledge distillation to multi-layer, multi-scale scheme, enlarging the distillation gap between normal samples and anomalies. PaDiM~\cite{padim} models normal distribution using pre-trained features, then utilize a distance metric to measure the anomalies.  Wang et al.~\cite{local-global} compare the embeddings of local pattern and global pattern to detect anomalies. 

\textbf{Transformer in anomaly detection.} Transformer~\cite{vaswani2017attention} has been successfully used in computer vision~\cite{carion2020end}. 
Some attempts also try to utilize transformer for anomaly detection. InTra~\cite{intra} adopts transformer to recover the image by recovering all masked patches one by one. VT-ADL~\cite{vtadl} and AnoVit~\cite{anovit} both apply transformer encoder to reconstruct images. However, these methods mainly focus on indistinguishable raw pixels, and do not figure out why transformer brings improvement. In contrast, we reconstruct pre-trained features instead of raw pixels. We also confirm the efficacy of the query embedding in attention layer to prevent the ``identical shortcut''. 

\section{Method}\label{sec:method}

In this part, we first introduce the architecture of ADTR, followed by the analysis of why transformer could limit to reconstruct anomalies well. Finally, we propose two loss functions to extend our approach compatible with available anomalies.

\begin{figure*}[tb]
    \centering
    \includegraphics[width=0.95\linewidth]{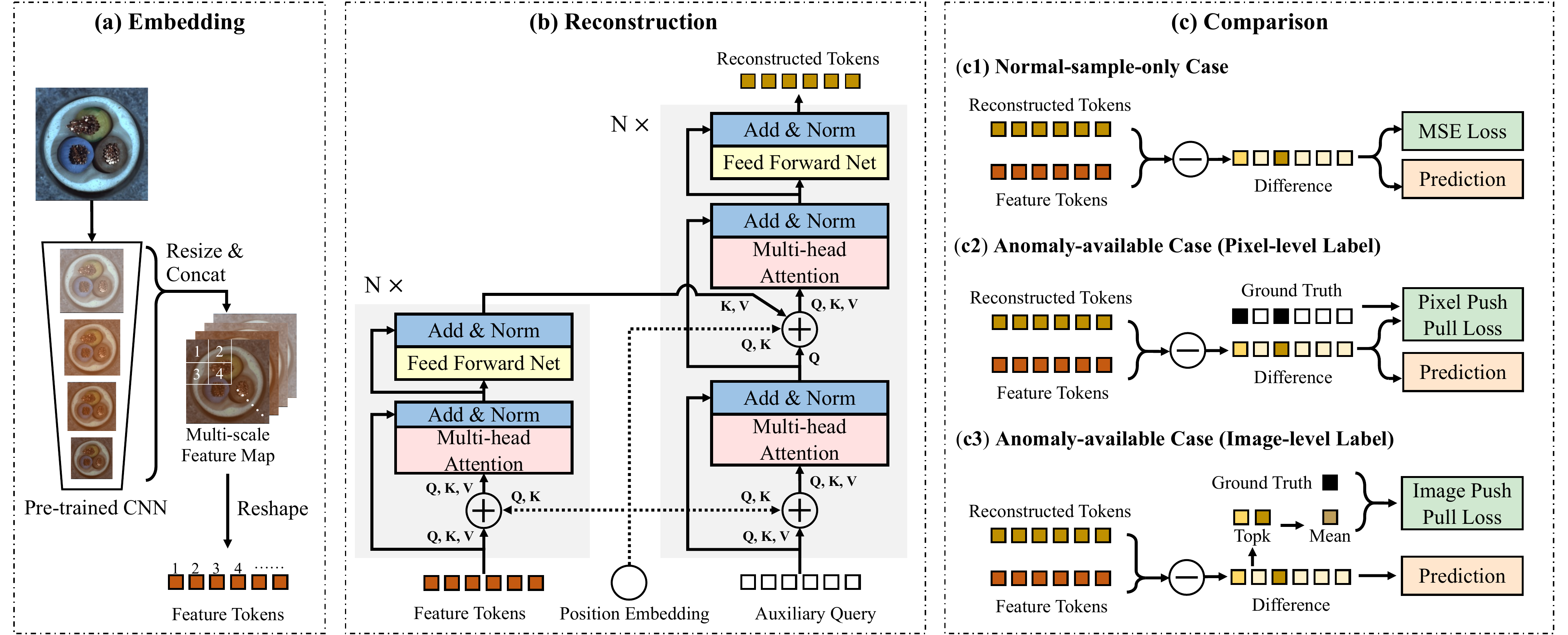}
    \vspace{-6pt}
    \caption{\textbf{Overview of our method.} (a) Embedding: a pre-trained CNN backbone is applied to extract the multi-scale features. (b) Reconstruction: a transformer is utilized to reconstruct the feature tokens with an auxiliary learnable query embedding. (c) Comparison: our approach is compatible with both normal-sample-only case and anomaly-available case. The anomaly score maps are obtained through the differences between extracted and reconstructed features. 
}
\label{fig:main}
\vspace{-16pt}
\end{figure*}


\subsection{Architecture}

\textbf{Embedding}. 
A frozen pre-trained CNN backbone is first utilized for feature extraction (\cref{fig:main}a). Here we use EfficientNet-B4~\cite{efficientnet} pre-trained on ImageNet. The features from \textit{layer1} to \textit{layer5} are resized to the same size, then concatenated together to form a multi-scale feature map, $\bm{f} \in \mathbb{R}^{C\times H\times W}$. Note that here we define \textit{layer} as the combination of stages with the same size of features. We adopt multi-scale feature map because feature maps from different layers have different levels of receptive fields thus are sensitive to different anomalies. 

\textbf{Reconstruction}. 
The reconstruction stage is shown in \cref{fig:main}b. The feature map, $\bm{f} \in \mathbb{R}^{C\times H\times W}$, is first split to $H \times W$ feature tokens. To reduce the computation consumption, a $1\times1$ convolution is applied to reduce the dimension of these tokens before they are fed into the transformer. Also, their dimensions are recovered by another $1\times1$ convolution when output by transformer. 
The transformer encoder embeds the input feature tokens into a latent feature space. Each encoder layer follows the standard architecture~\cite{vaswani2017attention} with multi-head attention, feed forward network (FFN), residual connection, and normalization. 
%
The transformer decoder follows the standard architecture~\cite{vaswani2017attention} with an auxiliary query embedding. The auxiliary query is a learned embedding with the same size of the input feature tokens. The transformer decoder transforms this learned query embedding to reconstruct the feature tokens using multi-head self-attention and encoder-decoder attention mechanisms. The learned position embedding~\cite{carion2020end} is included because transformer is permutation-invariant. 

\textbf{Comparison}. 
In normal-sample-only case, the model is trained with the MSE loss, $\mathcal{L}_{norm}$, between the backbone extracted features, $\bm{f}$, and the reconstructed features, $\hat{\bm{f}} \in \mathbb{R}^{C\times H\times W}$, as follows,
\begin{equation}\label{equ:mse_loss}
\mathcal{L}_{norm} = \frac{1}{H \times W} ||\bm{f} - \hat{\bm{f}}||^2_2. 
\end{equation}

\textbf{Inference}. We first define the feature difference map, $\bm{d}(i,u)$, as, 
\begin{equation}\label{equ:diff_feat}
\bm{d}(i,u) = \bm{f}(i,u) - \hat{\bm{f}}(i,u), 
\end{equation}
where $i$ represents the index of channel, $u$ is the index of spatial position (height together with width for simplicity). 
\textit{Anomaly localization} aims to localize anomalous regions, producing an anomaly score map, $\bm{s}(u)$, which assigns an anomaly score for each pixel, $u$. $\bm{s}(u)$ is calculated as the $L2$ norm of the feature difference vector, $\bm{d}(:, u)$. 
\begin{equation}\label{equ:score_map}
    \bm{s}(u) = ||\bm{d}(:, u)||_{2}.
\end{equation}
\textit{Anomaly detection} aims to detect whether an image contains anomalous regions. We intuitively take the maximum value of the averagely pooled $\bm{s}(u)$ as the anomaly score of the whole image. 


\subsection{Preventing ``Identical Mapping'' with Transformer}\label{subsec:prevent}

We suspect that, compared with CNN, the query embedding in attention layer makes transformer difficult to learn an ``identical mapping''. We denote the features in a normal image as $\bm{x}^+ \in \mathbb{R}^{K \times C}$, where $K$ is the feature number, $C$ is the channel dimension. The features in an anomalous image are denoted as $\bm{x}^- \in \mathbb{R}^{K \times C}$. We take a 1-layer network as the reconstruction net, which is trained on $\bm{x}^+$ with the MSE loss and tested to detect anomalous regions in $\bm{x}^-$. 

\textbf{Convolutional layer in CNN}. We first visit a fully-connected layer, whose weights and bias are denoted as $\bm{w} \in \mathbb{R}^{C \times C}, \bm{b} \in \mathbb{R}^{C}$, respectively. When using this layer as the reconstruction model of normal samples, it can be written as, 
\begin{equation}
    \bm{\hat{x}} = \bm{x}^+ \bm{w} + \bm{b} \in \mathbb{R}^{K \times C}.
\end{equation}
With the MSE loss pushing $\bm{\hat{x}}$ to $\bm{x}^{+}$, the model may take shortcut to regress $\bm{w} \to \bm{I}$ (identity matrix), $\bm{b} \to \bm{0}$. Ultimately, this model could also reconstruct $\bm{x}^{-}$ well, failing in anomaly detection. A convolutional layer with 1$\times$1 kernel is equivalent to a fully-connected layer. Besides, An $n \times n$ ($n>1$) kernel has more parameters and larger capacity, and can complete whatever 1$\times$1 kernel can. Thus, the convolutional layer also has the chance to learn a shortcut.

\textbf{Transformer with query embedding} contains an attention layer with a learnable query embedding, $\bm{q} \in \mathbb{R}^{K \times C}$. This attention layer can be denoted as, 
\begin{equation}
    \bm{\hat{x}} = \mathtt{softmax}(\bm{q} (\bm{x}^{+})^T / \sqrt{C}) \bm{x}^+ \in \mathbb{R}^{K \times C}.
\end{equation}
To push $\bm{\hat{x}}$ to $\bm{x}^{+}$, the attention map, $\mathtt{softmax}(\bm{q} (\bm{x}^{+})^T / \sqrt{C})$, should approximate $\bm{I}$ (identity matrix), so $\bm{q}$ must be highly related to $\bm{x}^{+}$. Considering that $\bm{q}$ in the trained model is relevant to normal samples, the model could not reconstruct $\bm{x}^{-}$ well. The ablation study in \cref{subsec:ablation} shows that without the attention layer or the query embedding, the performance of transformer respectively drops by 2.4\% or 3\%, which is almost the same as CNN. This reflects that the query embedding in attention layer helps prevent transformer from learning an ``identical shortcut''.

\subsection{Adaptation with Anomaly-available Case} \label{subsec:adaptation}

In practice, anomalies gradually increase with the runs of production lines, which brings the demands of compatibility with these increasing anomalies. Thus we adapt ADTR to ADTR+ for compatibility with the anomaly-available case. 

\textbf{Adaptation with pixel-level labels}. Inspired by~\cite{liznerski2020explainable}, we firstly calculate a pseudo-Huber loss, $\bm{\phi}(u)$, using the feature difference map, $\bm{d}(i,u)$. 
\begin{equation}\label{equ:Huber}
    \bm{\phi}(u) = ((\frac{1}{C}\sum_{i}^C |\bm{d}(i,u)|)^2 + 1)^{\frac{1}{2}} - 1.
\end{equation}
The pseudo-Huber loss, $\bm{\phi}(u)$, is designed as a difference map, which is easy to train and extend. Then the reconstruction loss function with pixel-level labels is denoted as $\mathcal{L}_{px}$ and could be described as a ``push-pull loss'' as, 
\begin{equation}\label{equ:loss_px}
\mathcal{L}_{px} = \frac{1}{HW}\sum_u^{HW} (1-\bm{y}(u))\bm{\phi}(u) - \alpha \log(1-\exp(-\frac{1}{H W}\sum_u^{HW} \bm{y}(u) \bm{\phi}(u))),
\end{equation}
where the first term pulls the reconstructed normal features to the extracted features, and the second term pushes the reconstructed anomalous features away from the original features, $\bm{y}(u)$ is the pixel-level label (0 for normal sample and 1 for anomaly) and $\alpha$ is a weight term. 

\textbf{Adaptation with image-level labels}. Since anomaly samples could contain both anomalous and normal regions, simply treating all regions of anomaly samples as anomalous regions confuses the model. Considering that larger values in $\bm{\phi}(u)$ are more likely to be anomalous regions, we firstly collect $k$ maximum values of $\bm{\phi}(u)$, then calculate their mean as the anomaly score of the image. 
\begin{equation}
    q = \frac{1}{k} \sum \mathtt{top\_k}(\bm{\phi}). 
\end{equation}
Then the image-level loss, $\mathcal{L}_{img}$, could be calculated as,
\begin{equation}\label{eq_image_loss}
\begin{aligned}
    \mathcal{L}_{img} = (1-y)q - \alpha y\log(1-\exp(-q)),
\end{aligned}
\end{equation}
where $y$ is the image-level label (0 for normal sample and 1 for anomaly) and $\alpha$ is a weight term. In $\mathcal{L}_{img}$, the first term pulls the reconstructed features of normal samples towards the extracted features, while the second term pushes the reconstructed features of anomalies away from the extracted features. 

\section{Experiment}\label{sec:Experiment}

\subsection{Dataset}

\begin{wrapfigure}{r}{0.5\linewidth}
\centering
\vspace{-30pt}
\includegraphics[width=1\linewidth]{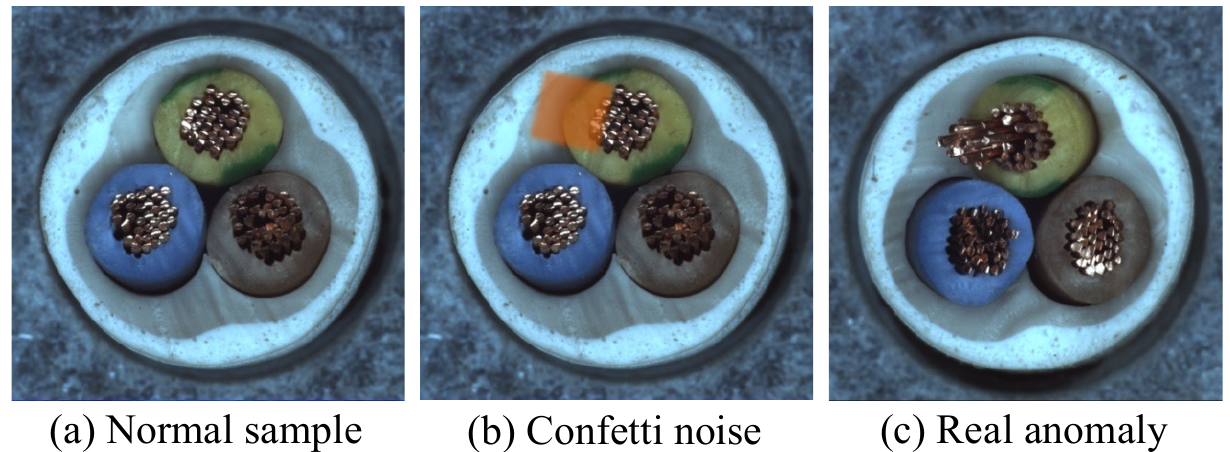}
\vspace{-18pt}
\caption{\textbf{Synthetic anomalies} by adding confetti noise on normal samples.}
\label{fig:aran}
\vspace{-30pt}
\end{wrapfigure}
\textbf{MVTec-AD} \cite{bergmann2019mvtec} is a multi-category, multi-defect, industrial anomaly detection dataset with 15 categories. The ground-truth includes both image labels and anomaly segmentation. In \textit{normal-sample-only case}, we follow the original setting to use normal samples for training, and test on both normal and anomaly samples. In \textit{anomaly-available case}, following~\cite{liznerski2020explainable}, we synthesize anomalies by adding confetti noise on normal samples (\cref{fig:aran}).

\textbf{CIFAR-10}~\cite{krizhevsky2009learning} is a classical classification dataset with 10 classes. Each class has 5000 images for training and 1000 images for testing. In \textit{normal-sample-only case}, following~\cite{gradcon}, the training set of one class is used for training, and the test set contains normal images of the same class and the same number of anomaly images randomly sampled from other classes. In \textit{anomaly-available case}, an irrelevant dataset, CIFAR-100~\cite{krizhevsky2009learning}, is used as an auxiliary dataset. We randomly select the same number of images from CIFAR-100 as anomalies.


\subsection{Anomaly Detection on MVTec-AD} \label{subsec:mvtec_pixel}

The performance of our method is evaluated on anomaly detection and localization tasks of MVTec-AD~\cite{bergmann2019mvtec}. 

\textbf{Setup}. 
The sizes of the image and feature map are selected as $256 \times 256$ and $16 \times 16$, respectively. The numbers of the encoder layer and decoder layer ($N$ in \cref{fig:main}) in transformer are both set as 4. The features from \textit{layer1} to \textit{layer5} of EfficientNet-B4~\cite{efficientnet} are resized and concatenated to form a 720-channel feature map. The reduced channel dimension is set as 256. AdamW optimizer~\cite{AdamW} with weight decay $1 \times 10^{-4}$ is used for training with batch size 16. In \textit{normal-sample-only case}, models are trained with $\mathcal{L}_{norm}$ in Eq. (\ref{equ:mse_loss}) for 500 epochs. The learning rate is $1 \times 10^{-4}$ initially, and dropped by 0.1 after 400 epochs. In \textit{anomaly-available case}, the pixel-level loss, $\mathcal{L}_{px}$, in Eq. (\ref{equ:loss_px}) is adopted for training, where $\alpha$ is chosen as 0.003. The trained model in normal-sample-only case is firstly loaded. Then the model is trained for 300 epochs with the learning rate of $1 \times 10^{-4}$ for first 200 epochs and $1 \times 10^{-5}$ for last 100 epochs.  

\begin{figure}[tb]
    \centering
    \includegraphics[width=0.9\linewidth]{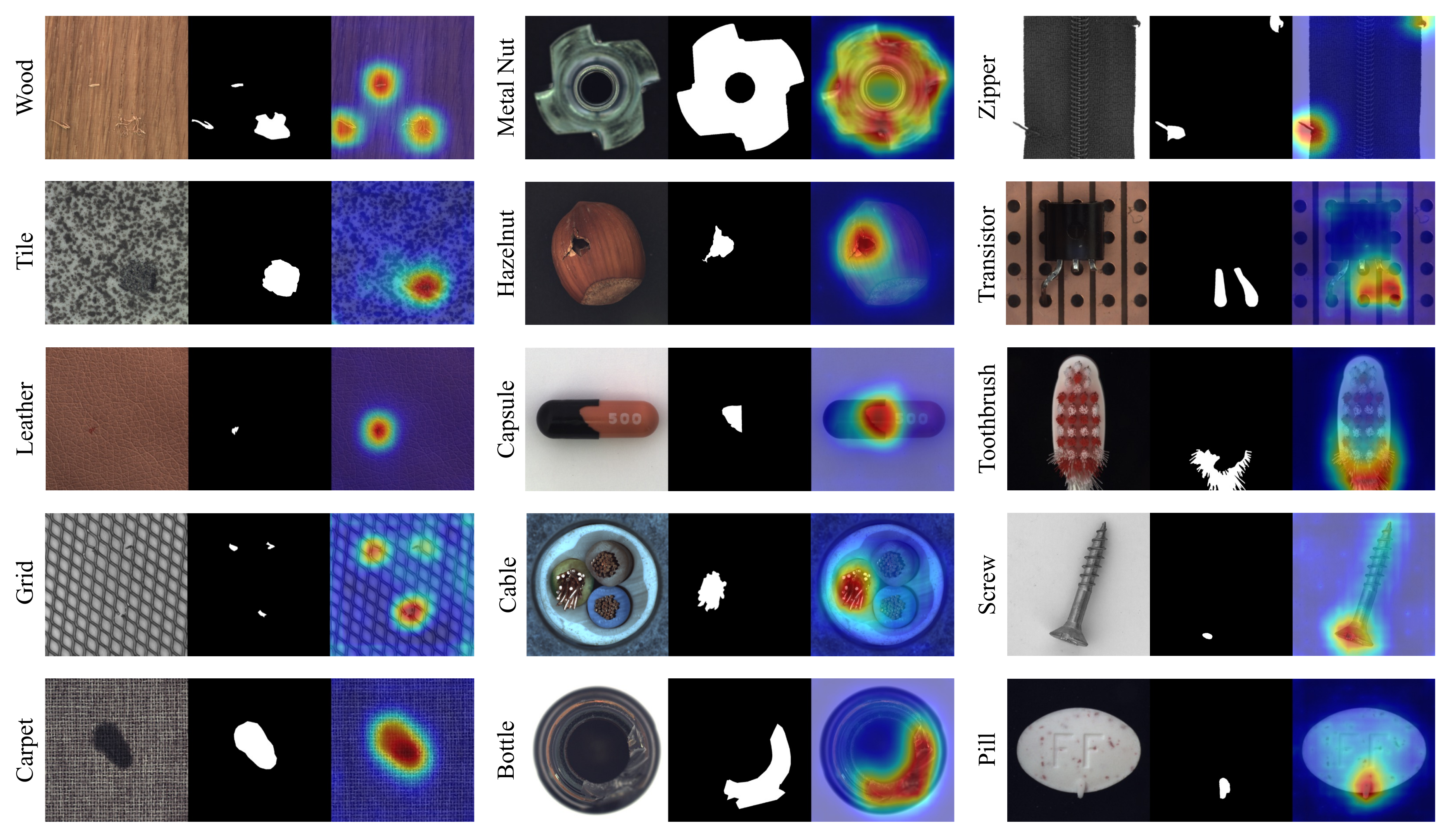}
    \vspace{-10pt}
    \caption{\textbf{Anomaly detection results on MVTec-AD~\cite{bergmann2019mvtec}}. From left to right: the anomaly sample, the ground-truth, and the anomaly score map of ADTR.}
    \label{fig:mvtecresult}
    \vspace{-18pt}
\end{figure}

\textbf{Qualitative results} on MVTec-AD are shown in \cref{fig:mvtecresult}. Our approach successfully detects different kinds of anomalies with high localization accuracy. Especially, for the shown ``Metal Nut'' example, where the anomaly is a flipped normal sample, our approach detects the ``flip'' anomaly successfully though there are no obvious vision anomalies like texture disorder nor color change. 

\textbf{Quantitative results of anomaly localization} are given in \cref{tab:pixel_auc}. Our approach is compared with AnoGAN \cite{anogan}, SCADN \cite{scadn}, SSIM-AE \cite{ssimae}, VEVAE \cite{vevae}, SMAI \cite{smai}, VAE-Grad \cite{vae-grad}, P-Net \cite{pnet}, KDAD \cite{kdad}, Loc-Glo \cite{local-global}, FCDD \cite{liznerski2020explainable},  PSVDD \cite{yi2020patch}, SPADE \cite{spade}. With pure normal samples, ADTR stably outperforms the best baseline, SPADE \cite{spade}, by 1.2\%. With merely simple synthetic anomalies, the performance of ADTR+ is further improved by 0.3\%.

\begin{table*}[tb]
  \setlength\tabcolsep{1.5pt}
  \centering
  \scriptsize
  \caption{\textbf{Anomaly localization results under pixel-level AUROC metric on MVTec-AD~\cite{bergmann2019mvtec}.}}
    \begin{tabular}{c|ccccc|cccccccccc|c}
    \toprule
    & \multicolumn{5}{c}{Texture}  & \multicolumn{10}{|c|}{Object} & \\
    \midrule
    & \rotatebox[]{270}{Carp.} & \rotatebox[]{270}{Grid}  & \rotatebox[]{270}{Leat.} & \rotatebox[]{270}{Tile} & \rotatebox[]{270}{Wood} & \rotatebox[]{270}{Bott.} & \rotatebox[]{270}{Cable} & \rotatebox[]{270}{Caps.} & \rotatebox[]{270}{Haze.} & \rotatebox[]{270}{Meta.} & \rotatebox[]{270}{Pill} & \rotatebox[]{270}{Screw} & \rotatebox[]{270}{Toot.} & \rotatebox[]{270}{Tran.} & \rotatebox[]{270}{Zipp.} & \rotatebox[]{270}{Mean} \\
    \midrule
    AnoGAN \cite{anogan} & 54   & 58  & 64    & 50  & 62  & 86   & 78  & 84    & 87     & 76      & 87  & 80  & 90       & 80       & 78   & 74  \\
    SCADN \cite{scadn} & 64.9 & 79.6 & 76.3 & 67.7 & 67.2 & 69.6 & 81.4 & 68.7 & 88.4 & 75.4 & 74.7 & 87.6 & 90.1 & 68.9 & 67.0 & 75.2 \\
    SSIM-AE \cite{ssimae} & 87   & 94  & 78    & 59  & 73  & 93   & 82  & 94    & 97     & 89      & 91  & 96  & 92       & 90       & 88   & 86  \\
    VEVAE \cite{vevae} & 78   & 73  & 95    & 80  & 77  & 87   & 90    & 74    & 98     & 94      & 83  & 97  & 94       & 93       & 78   & 86  \\
    SMAI \cite{smai} & 88   & 97  & 86    & 62  & 80  & 86   & 92  & 93    & 97     & 92      & 92  & 96  & 96       & 85       & 90   & 89  \\
    VAE-Grad \cite{vae-grad} & 74   & 96  & 93    & 65  & 84  & 92   & 91  & 92    & 98     & 91      & 93  & 95  & 99       & 92       & 87   & 89  \\
    P-Net \cite{pnet} & 57   & \textbf{98}  & 89    & \textbf{97}  & \textbf{98}  & \textbf{99}   & 70  & 84    & 97     & 79      & 91  & \textbf{100}  & 99       & 82       & 90   & 89  \\
    KDAD \cite{kdad} & 95.6  & 91.8 & 98.1 & 82.8 & 84.8 & 96.3 & 82.4 & 95.9 & 94.6 & 86.4 & 89.6 & 96.0 & 96.1 & 76.5 & 93.9 & 90.7 \\
    Loc-Glo \cite{local-global} & 96  & 78 & 90 & 80 & 81 & 93 & 94 & 90 & 84 & 91 & 93 & 96 & 96 & \textbf{100} & \textbf{99} & 91 \\
    FCDD \cite{liznerski2020explainable} & 96   & 91  & 98    & 91  & 88  & 97   & 90  & 93    & 95     & 94      & 81  & 86  & 94       & 88       & 92   & 92  \\
    PSVDD \cite{yi2020patch} & 92.6  & 96.2 & 97.4 & 91.4 & 90.8 & 98.1  & 96.8 & 95.8 & 97.5 & 98.0 & 95.1 & 95.7 & 98.1 & 97.0 & 95.1 & 95.7 \\
    SPADE \cite{spade} & 97.5 & 93.7 & 97.6 & 87.4 & 88.5 & 98.4 & \textbf{97.2} & 99.0 & \textbf{99.1} & \textbf{98.1} & 96.5 & 98.9 & 97.9 & 94.1 & 96.5 & 96.0 \\
    \midrule
    ADTR(ours) & 98.7 & 95.0 & 98.1 & 93.8 & 91.2 & 98.0 & 96.8 & \textbf{99.1} & 98.6 & 97.0 & 98.3 & 99.3 & 98.5 & 97.9 & 97.2 & 97.2 \\
    ADTR+(ours) & \textbf{98.8} & 94.2 & \textbf{98.6} & 95.9 & 93.0 & 98.0 & 97.0 & \textbf{99.1} & 98.8 & 96.8 & \textbf{98.7} & 99.3 & \textbf{99.2} & 97.8 & 97.6 & \textbf{97.5} \\
    \bottomrule
    \end{tabular}
  \label{tab:pixel_auc}
  \vspace{-15pt}
\end{table*}

\textbf{Quantitative results of anomaly detection} are shown in \cref{tab:img_auc}. Our approach is compared with GANomaly \cite{ganomaly}, SCADN \cite{scadn}, ARNet \cite{arnet}, SPADE \cite{spade}, KDAD \cite{kdad}, PSVDD \cite{yi2020patch}, TS \cite{ts}. ADTR considerably exceeds all baseline methods ($\ge$ 3.9\%) with only normal samples. The performance of ADTR+ is improved by 0.5\% with simple synthetic anomalies.

\begin{table*}[tb]
  \setlength\tabcolsep{1.5pt}
  \centering
  \scriptsize
  \caption{\textbf{Anomaly detection results under image-level AUROC metric on MVTec-AD~\cite{bergmann2019mvtec}.}}
  \begin{tabular}{c|ccccc|cccccccccc|c}
    \toprule
    & \multicolumn{5}{c}{Texture}  & \multicolumn{10}{|c|}{Object} & \\
    \midrule
    & \rotatebox[]{270}{Carp.} & \rotatebox[]{270}{Grid}  & \rotatebox[]{270}{Leat.} & \rotatebox[]{270}{Tile} & \rotatebox[]{270}{Wood} & \rotatebox[]{270}{Bott.} & \rotatebox[]{270}{Cable} & \rotatebox[]{270}{Caps.} & \rotatebox[]{270}{Haze.} & \rotatebox[]{270}{Meta.} & \rotatebox[]{270}{Pill} & \rotatebox[]{270}{Screw} & \rotatebox[]{270}{Toot.} & \rotatebox[]{270}{Tran.} & \rotatebox[]{270}{Zipp.} & \rotatebox[]{270}{Mean} \\
    \midrule
    GANomaly \cite{ganomaly} & 69.9 & 70.8 & 84.2 & 79.4 & 83.4 & 89.2 & 75.7 & 73.2 & 78.5 & 70.0 & 74.3 & 74.6 & 65.3 & 79.2 & 74.5 & 76.2 \\
    SCADN \cite{scadn}  & 50.4 & 98.3 & 65.9 & 79.2 & 96.8 & 95.7 & 85.6 & 76.5 & 83.3 & 62.4 & 81.4 & 83.1 & 98.1 & 86.3 & 84.6 & 81.8 \\
    ARNet \cite{arnet} & 70.6 & 88.3 & 86.2 & 73.5 & 92.3 & 94.1 & 83.2 & 68.1 & 85.5 & 66.7 & 78.6 & \textbf{100} & \textbf{100} & 84.3 & 87.6 & 83.9 \\
    SPADE \cite{spade} & - & - & - & - & - & - & - & - & - & - & - & - & - & - & - & 85.5 \\
    KDAD \cite{kdad} & 79.3 & 78.0 & 95.1 & 91.6 & 94.3 & 99.4 & 89.2 & 80.5 & 98.4 & 73.6 & 82.7 & 83.3 & 92.2 & 85.6 & 93.2 & 87.7 \\
    PSVDD \cite{yi2020patch} & 98.6 & 90.3 & 76.7 & 92.9 & 94.6 & 92.0 & 90.9 & \textbf{94.0} & 86.1 & 81.3 & \textbf{97.8} & \textbf{100} & 91.5 & 96.5 & \textbf{97.9} & 92.1 \\
    TS \cite{ts} & 95.3 & \textbf{98.7} & 93.4 & 95.8 & 95.5 & 96.7 & 82.3 & 92.8 & 91.4 & 94.0 & 86.7 & 87.4 & 98.6 & 83.6 & 95.8 & 92.5 \\
    \midrule
    ADTR(ours) & \textbf{100}  & 97.5 & \textbf{100} & \textbf{100} & \textbf{99.9} & \textbf{100} & \textbf{92.5} & 93.1 & \textbf{100} & \textbf{94.9} & 92.1 & 94.0 & 93.1 & 97.6 & 95.8 & 96.4 \\
    ADTR+(ours) & \textbf{100} & 97.8 & \textbf{100} & \textbf{100} & \textbf{99.9} & \textbf{100} & \textbf{92.5} & 92.5 & 99.9 & 94.5 & 93.3 & 94.2 & 93.9 & \textbf{98.0} & 97.0 & \textbf{96.9} \\
    \bottomrule
    \end{tabular}
  \label{tab:img_auc}
  \vspace{-15pt}
\end{table*}

\subsection{Anomaly Detection on CIFAR-10}\label{subsec:cifar10_pixel}

\begin{table*}[tb]
  \centering
  \scriptsize
  \caption{\textbf{Anomaly detection results under image-level AUROC metric on CIFAR-10 \cite{krizhevsky2009learning}.}}
    \begin{tabular}{c|cccccccccc|c}
    \toprule
     & Airplane & Automobile & Bird & Cat & Deer & Dog & Frog & Horse & Ship & Truck & Mean \\
     \midrule
     VAE \cite{vae} & 63.4 & 44.2 & 64.0 & 49.7 & 74.3 & 51.5 & 74.5 & 52.7 & 67.4 & 41.6 & 58.3 \\
     KDE \cite{kde} & 65.8 & 52.0 & 65.7 & 49.7 & 72.7 & 49.6 & 75.8 & 56.4 & 68.0 & 54.0 & 61.0 \\
     AnoGAN \cite{anogan} & 67.1 & 54.7 & 52.9 & 54.5 & 65.1 & 60.3 & 58.5 & 62.5 & 75.8 & 66.5 & 61.8 \\
     LSA \cite{lsa} & 73.5 & 58.0 & 69.0 & 54.2 & 76.1 & 54.6 & 75.1 & 53.5 & 71.7 & 54.8 & 64.1 \\
     DSVDD \cite{svdd} & 61.7 & 65.9 & 50.8 & 59.1 & 60.9 & 65.7 & 67.7 & 67.3 & 75.9 & 73.1 & 64.8 \\
     OCGAN \cite{ocgan} & 75.7 & 53.1 & 64.0 & 62.0 & 72.3 & 62.0 & 72.3 & 57.5 & 82.0 & 55.4 & 65.7 \\
     GradCon \cite{gradcon} & 76.0 & 59.8 & 64.8 & 58.6 & 73.3 & 60.3 & 68.4 & 56.7 & 78.4 & 67.8 & 66.4 \\
     Loc-Glo~\cite{local-global} & 79.1 & 70.3 & 67.5 & 56.1 & 73.9 & 63.8 & 73.2 & 67.4 & 81.4 & 72.2 & 70.5 \\
     TS \cite{ts} & 78.9 & 84.9 & 73.4 & 74.8 & 85.1 & 79.3 & 89.2 & 83.0 & 86.2 & 84.8 & 82.0 \\
     GT \cite{gt} & 76.2 & 84.8 & 77.1 & 73.2 & 82.8 & 84.8 & 82.0 & 88.7 & 89.5 & 83.4 & 82.3 \\
     KDAD \cite{kdad} & 90.5 & 90.4 & 80.0 & 77.0 & 86.7 & 91.4 & 89.0 & 86.8 & 91.5 & 88.9 & 87.2 \\
     \midrule
     ADTR(ours) & 94.1 & 97.4 & 92.3 & 89.0 & 93.2 & 94.4 & 97.4 & 95.8 & 96.3 & 96.7 & 94.7 \\
     ADTR+(ours) & \textbf{96.2} & \textbf{98.0} & \textbf{94.5} & \textbf{91.7} & \textbf{95.1} & \textbf{95.6} &  \textbf{98.0} & \textbf{97.1} & \textbf{98.0} &  \textbf{96.9} & \textbf{96.1} \\
     \toprule
    \end{tabular}
  \label{tab:cifar}
  \vspace{-15pt}
\end{table*}

To further validate the anomaly detection ability, we evaluate our model in the unsupervised one-class classification task of CIFAR-10~\cite{krizhevsky2009learning}. 

\textbf{Setup}. The setup is the same as that in \cref{subsec:mvtec_pixel} except the followings. First, the sizes of the image and feature map are $32 \times 32$ and $8 \times 8$, respectively. Second, in anomaly-available case, the model is trained with the image-level loss, $\mathcal{L}_{img}$, in Eq. (\ref{eq_image_loss}), where $\alpha$ and $k$ are selected as 0.003 and 20, respectively.


\textbf{Quantitative results on CIFAR-10} are shown in \cref{tab:cifar}. The competitors include: VAE~\cite{vae}, KDE~\cite{kde}, AnoGAN~\cite{anogan}, LSA~\cite{lsa}, DSVDD \cite{svdd}, OCGAN~\cite{ocgan}, GradCon~\cite{gradcon}, Loc-Glo~\cite{local-global}, TS~\cite{ts}, GT~\cite{gt},  KDAD~\cite{kdad}. ADTR surpasses KDAD \cite{kdad} by a great margin (7.5\%) when training in normal-sample-only case. In anomaly-available case, the performance of ADTR+ is further improved by 1.4\% with the help of external irrelevant dataset, reflecting the effectiveness of the designed image-level loss function,  $\mathcal{L}_{img}$.




\subsection{Ablation Study}\label{subsec:ablation}

Extensive ablation studies with pixel-level AUROC metric are conducted on anomaly localization task of MVTec-AD \cite{bergmann2019mvtec}. 

\textbf{Attention and auxiliary query embedding}. As shown in \cref{tab:ablation}a, a CNN revised from ResNet~\cite{he2016deep} is firstly included as the baseline of the reconstruction model. (1) The replacement of the attention layer is a concatenation followed by projection. If we remove the attention layer (w/o Attn) from the transformer, the performance shows no obvious superiority to CNN. (2) Without the auxiliary query embedding (w/o Query), meaning that only the encoder embedding is input to the decoder, the performance is even worse than CNN. (3) Equipped with both attention and auxiliary query embedding (Attn+Query), transformer stably outperforms CNN by 2.8\%. This proves our assertion in \cref{subsec:prevent} that the auxiliary query embedding in attention layer helps prevent transformer from reconstructing anomalies well. 

\textbf{Reconstructed target}. In \cref{tab:ablation}b, reconstructing features surpasses pixel values substantially, indicating that the features extracted by pre-trained backbone are more distinguishable for normal samples and anomalies than raw pixels. 

\textbf{Backbone and multi-scale features}. (1) As shown in \cref{tab:ablation}c, four different backbones all achieve quite good performance, reflecting that our method could cooperate with different types of backbones. (2) In \cref{tab:ablation}d, multi-scale features obviously outperform last-layer feature, because multi-scale features contain different levels of receptive fields thus are sensitive to different anomalies.

\begin{table}[t]
  \centering
  \scriptsize
  \caption{\textbf{Ablation study} on (a) attention \& auxiliary query embedding, (b) reconstructing pixels \textit{vs.} features, (c) backbone, and (d) multi-scale features under pixel-level AUROC metric on anomaly localization of MVTec-AD~\cite{bergmann2019mvtec}.}
  \vspace{5pt}
  \begin{minipage}[t]{0.58\textwidth}
    \centering
    (a) Attention \& auxiliary query embedding
    \begin{threeparttable}
    \begin{tabular}{ccccc}
    \toprule
    & CNN & w/o Attn & w/o Query & Attn+Query \\
    \midrule
    Pixel AUROC & 94.4 & 94.8 & 94.2 & \textbf{97.2} \\
    \bottomrule
    \end{tabular}
    \end{threeparttable}
    \\
    \vspace{3pt}
    (c) Backbone
    \begin{threeparttable}
    \begin{tabular}{ccccc}
    \toprule
    & Res-18 & Res-34 & Efficient-B0 & Efficient-B4 \\
    \midrule
    Pixel AUROC & 95.3 & 95.7 & 96.4 & \textbf{97.2} \\
    \bottomrule
    \end{tabular}
    \end{threeparttable}
  \end{minipage}
  \begin{minipage}[t]{0.4\textwidth}
    \centering
    (b) Reconstructing pixels \textit{vs.} features
    \begin{threeparttable}
    \begin{tabular}{ccccc}
    \toprule
    & Pixels & Features \\
    \midrule
    Pixel AUROC  & 91.3 & \textbf{97.2} \\
    \bottomrule
    \end{tabular}
    \end{threeparttable}
    \\
    \vspace{5pt}
    (d) Multi-scale features
    \begin{threeparttable}
    \begin{tabular}{ccccc}
    \toprule
    & Last-layer & Multi-scale \\
    \midrule
    Pixel AUROC  & 96.0 & \textbf{97.2} \\
    \bottomrule
    \end{tabular}
    \end{threeparttable}
  \end{minipage}
  \label{tab:ablation}
  \vspace{-16pt}
\end{table}

\subsection{Visualization of Feature Difference Vectors}

\begin{wrapfigure}{r}{0.6\linewidth}
\centering
\vspace{-20pt}
\includegraphics[width=1\linewidth]{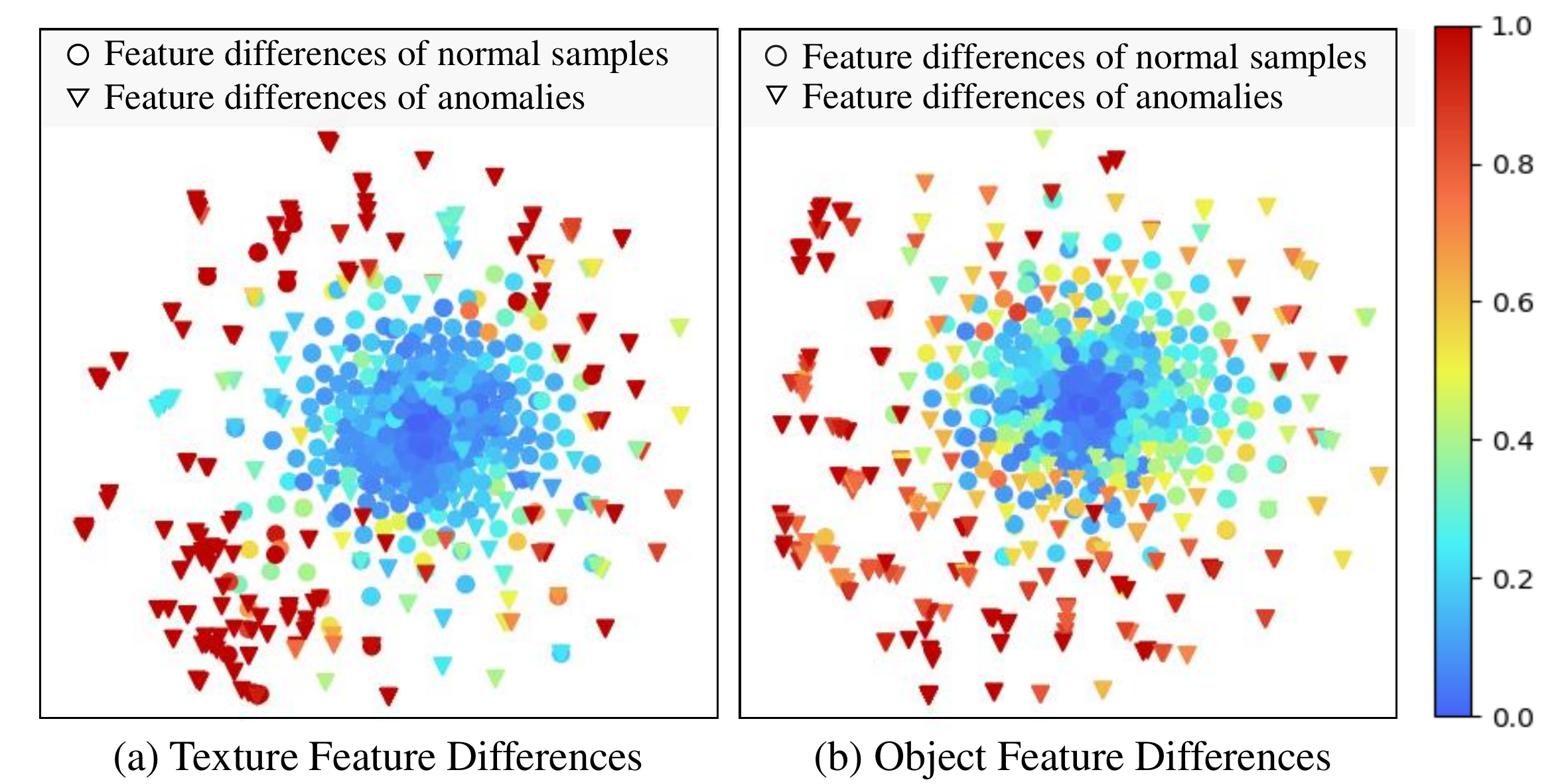}
\vspace{-20pt}
\caption{\textbf{visualization of feature difference vectors} by t-SNE. Circles and triangles respectively represent normal samples and anomalies. The color map indicates the predicted anomaly possibility. Our method brings large generalization gap between normal samples and anomalies.}
\label{fig:vis_feat}
\vspace{-20pt}
\end{wrapfigure}
We visualize the feature difference vectors $\bm{d}(:,u)$ in Eq. (\ref{equ:diff_feat}) to better interpret our approach. Specifically, we randomly sample 600 feature difference vectors (normal : anomaly = 1:1) from MVTec-AD~\cite{bergmann2019mvtec}. Then t-SNE is utilized to visualize the high dimensional vectors in a 2D space, as shown in \cref{fig:vis_feat}. Firstly, normal samples and anomalies are mostly colored with blue and red, respectively, indicating good anomaly detection ability. Secondly, normal samples are well clustered, and there is a wide gap between the normal samples and anomalies. These observations indicate that our approach brings a large generalization gap between normal samples and anomalies.

\section{Conclusion}\label{sec:conclu}

In this paper, we propose anomaly detection transformer to utilize a transformer to reconstruct pre-trained features. First, the pre-trained features contain distinguishable semantic information. Second, the adoption of transformer prevents reconstructing anomalies well such that anomalies could be detected easily once the reconstruction fails. Our method brings a large generalization gap between normal samples and anomalies. Moreover, we propose novel loss functions to extend our approach from normal-sample-only case to anomaly-available case with both image-level labeled and pixel-level labeled anomalies, further improving the performance. Our approach achieves the state-of-the-art performance on anomaly detection benchmarks including MVTec-AD and CIFAR-10.

{
\small
\bibliographystyle{splncs04}
\bibliography{ref}

\begin{thebibliography}{10}
\providecommand{\url}[1]{\texttt{#1}}
\providecommand{\urlprefix}{URL }
\providecommand{\doi}[1]{https://doi.org/#1}

\bibitem{lsa}
Abati, D., Porrello, A., Calderara, S., Cucchiara, R.: Latent space
  autoregression for novelty detection. In: CVPR (2019)

\bibitem{ganomaly}
Akcay, S., Atapour-Abarghouei, A., Breckon, T.P.: {GAN}omaly: Semi-supervised
  anomaly detection via adversarial training. In: ACCV (2018)

\bibitem{vae}
An, J., Cho, S.: Variational autoencoder based anomaly detection using
  reconstruction probability. Special Lecture on IE  (2015)

\bibitem{bergmann2019mvtec}
Bergmann, P., Fauser, M., Sattlegger, D., Steger, C.: {MVTec AD:} a
  comprehensive real-world dataset for unsupervised anomaly detection. In: CVPR
  (2019)

\bibitem{ts}
Bergmann, P., Fauser, M., Sattlegger, D., Steger, C.: Uninformed students:
  Student-teacher anomaly detection with discriminative latent embeddings. In:
  CVPR (2020)

\bibitem{ssimae}
Bergmann, P., Lwe, S., Fauser, M., Sattlegger, D., Steger, C.: Improving
  unsupervised defect segmentation by applying structural similarity to
  autoencoders. In: International Conference on Computer Vision Theory and
  Applications (2019)

\bibitem{kde}
Bishop, C.M.: Pattern recognition and machine learning. springer (2006)

\bibitem{borghesi2019anomaly}
Borghesi, A., Bartolini, A., Lombardi, M., Milano, M., Benini, L.: Anomaly
  detection using autoencoders in high performance computing systems. In: AAAI
  (2019)

\bibitem{carion2020end}
Carion, N., Massa, F., Synnaeve, G., Usunier, N., Kirillov, A., Zagoruyko, S.:
  End-to-end object detection with transformers. In: ECCV (2020)

\bibitem{spade}
Cohen, N., Hoshen, Y.: Sub-image anomaly detection with deep pyramid
  correspondences. arXiv preprint arXiv:2005.02357  (2020)

\bibitem{padim}
Defard, T., Setkov, A., Loesch, A., Audigier, R.: {PaDim}: A patch distribution
  modeling framework for anomaly detection and localization. In: ICPR (2021)

\bibitem{vae-grad}
Dehaene, D., Frigo, O., Combrexelle, S., Eline, P.: Iterative energy-based
  projection on a normal data manifold for anomaly localization. In: ICLR
  (2019)

\bibitem{dehaene2020iterative}
Dehaene, D., Frigo, O., Combrexelle, S., Eline, P.: Iterative energy-based
  projection on a normal data manifold for anomaly localization. In: ICLR
  (2020)

\bibitem{arnet}
Fei, Y., Huang, C., Jinkun, C., Li, M., Zhang, Y., Lu, C.: Attribute
  restoration framework for anomaly detection. IEEE Transactions on Multimedia
  (2020)

\bibitem{gt}
Golan, I., El{-}Yaniv, R.: Deep anomaly detection using geometric
  transformations. In: Bengio, S., Wallach, H.M., Larochelle, H., Grauman, K.,
  Cesa{-}Bianchi, N., Garnett, R. (eds.) NIPS (2018)

\bibitem{gong2019memorizing}
Gong, D., Liu, L., Le, V., Saha, B., Mansour, M.R., Venkatesh, S., Hengel,
  A.v.d.: Memorizing normality to detect anomaly: Memory-augmented deep
  autoencoder for unsupervised anomaly detection. In: ICCV (2019)

\bibitem{he2016deep}
He, K., Zhang, X., Ren, S., Sun, J.: Deep residual learning for image
  recognition. In: CVPR (2016)

\bibitem{krizhevsky2009learning}
Krizhevsky, A.: Learning multiple layers of features from tiny images. Master's
  thesis, University of Tront  (2009)

\bibitem{gradcon}
Kwon, G., Prabhushankar, M., Temel, D., AlRegib, G.: Backpropagated gradient
  representations for anomaly detection. In: ECCV (2020)

\bibitem{smai}
Li, Z., Li, N., Jiang, K., Ma, Z., Wei, X., Hong, X., Gong, Y.: Superpixel
  masking and inpainting for self-supervised anomaly detection. In: BMVC (2020)

\bibitem{vevae}
Liu, W., Li, R., Zheng, M., Karanam, S., Wu, Z., Bhanu, B., Radke, R.J., Camps,
  O.: Towards visually explaining variational autoencoders. In: CVPR (2020)

\bibitem{liznerski2020explainable}
Liznerski, P., Ruff, L., Vandermeulen, R.A., Franks, B.J., Kloft, M.,
  M{\"{u}}ller, K.: Explainable deep one-class classification. In: ICLR (2021)

\bibitem{AdamW}
Loshchilov, I., Hutter, F.: Decoupled weight decay regularization. In: ICLR
  (2019)

\bibitem{vtadl}
Mishra, P., Verk, R., Fornasier, D., Piciarelli, C., Foresti, G.L.: {VT-ADL}: A
  vision transformer network for image anomaly detection and localization. In:
  International Symposium on Industrial Electronics (2021)

\bibitem{park2020learning}
Park, H., Noh, J., Ham, B.: Learning memory-guided normality for anomaly
  detection. In: CVPR (2020)

\bibitem{ocgan}
Perera, P., Nallapati, R., Xiang, B.: {OCGAN}: One-class novelty detection
  using {GANs} with constrained latent representations. In: CVPR (2019)

\bibitem{intra}
Pirnay, J., Chai, K.: Inpainting transformer for anomaly detection. arXiv
  preprint arXiv:2104.13897  (2021)

\bibitem{svdd}
Ruff, L., Vandermeulen, R., Goernitz, N., Deecke, L., Siddiqui, S.A., Binder,
  A., M{\"u}ller, E., Kloft, M.: Deep one-class classification. In: ICML (2018)

\bibitem{sabokrou2018adversarially}
Sabokrou, M., Khalooei, M., Fathy, M., Adeli, E.: Adversarially learned
  one-class classifier for novelty detection. In: CVPR (2018)

\bibitem{kdad}
Salehi, M., Sadjadi, N., Baselizadeh, S., Rohban, M.H., Rabiee, H.R.:
  Multiresolution knowledge distillation for anomaly detection. In: CVPR (2021)

\bibitem{anogan}
Schlegl, T., Seeb{\"o}ck, P., Waldstein, S.M., Schmidt-Erfurth, U., Langs, G.:
  Unsupervised anomaly detection with generative adversarial networks to guide
  marker discovery. In: International Conference on Information Processing in
  Medical Imaging (2017)

\bibitem{efficientnet}
Tan, M., Le, Q.: Efficientnet: Rethinking model scaling for convolutional
  neural networks. In: ICML (2019)

\bibitem{vaswani2017attention}
Vaswani, A., Shazeer, N., Parmar, N., Uszkoreit, J., Jones, L., Gomez, A.N.,
  Kaiser, {\L}., Polosukhin, I.: Attention is all you need. NIPS  (2017)

\bibitem{local-global}
Wang, S., Wu, L., Cui, L., Shen, Y.: Glancing at the patch: Anomaly
  localization with global and local feature comparison. In: CVPR (2021)

\bibitem{xia2020synthesize}
Xia, Y., Zhang, Y., Liu, F., Shen, W., Yuille, A.L.: Synthesize then compare:
  Detecting failures and anomalies for semantic segmentation. In: ECCV (2020)

\bibitem{scadn}
Yan, X., Zhang, H., Xu, X., Hu, X., Heng, P.A.: Learning semantic context from
  normal samples for unsupervised anomaly detection. In: AAAI (2021)

\bibitem{yi2020patch}
Yi, J., Yoon, S.: {Patch SVDD}: Patch-level {SVDD} for anomaly detection and
  segmentation. In: ACCV (2020)

\bibitem{anovit}
Yunseung, L., Pilsung, K.: {AnoViT}: Unsupervised anomaly detection and
  localization with vision transformer-based encoder-decoder. arXiv preprint
  arXiv:2203.10808  (2022)

\bibitem{zaheer2020old}
Zaheer, M.Z., Lee, J.h., Astrid, M., Lee, S.I.: Old is gold: Redefining the
  adversarially learned one-class classifier training paradigm. In: CVPR (2020)

\bibitem{pnet}
Zhou, K., Xiao, Y., Yang, J., Cheng, J., Liu, W., Luo, W., Gu, Z., Liu, J.,
  Gao, S.: Encoding structure-texture relation with p-net for anomaly detection
  in retinal images. In: ECCV (2020)

\end{thebibliography}
}

\appendix
\section*{Appendix}
\setcounter{figure}{0}
\renewcommand\thefigure{A\arabic{figure}}
\setcounter{table}{0}
\renewcommand\thetable{A\arabic{table}}
\setcounter{equation}{0}
\renewcommand\theequation{A\arabic{equation}}
\section{More Details}

\textbf{Backbone}. 
We use the ImageNet pre-trained EfficientNet-B4 \cite{efficientnet} \footnote{\scriptsize We use the EfficientNet-B4 checkpoint in \url{https://github.com/lukemelas/EfficientNet-PyTorch/releases/download/1.0/efficientnet-b4-6ed6700e.pth}} as the backbone. The features from layer1 to layer5 of EfficientNet-B4 \cite{efficientnet} have the channel of 24, 32, 56, 160, 448, respectively. Here we define ``layer'' as the combination of stages that have the same size of features. The 5 features are resized to the same size and concatenated together to form a 720-channel feature map. For MVTec-AD \cite{bergmann2019mvtec}, the image size and the feature size are set as $512 \times 512$ and $32 \times 32$, respectively. Therefore, a feature map with the shape of $32 \times 32 \times 720$ is obtained. For CIFAR-10 \cite{krizhevsky2009learning}, the image size is $32 \times 32$, which is quite small. Thus the size of the feature map is set relatively large (with the output stride of 4), so an $8 \times 8 \times 720$ feature map is obtained.

\textbf{Transformer}. 
A $1 \times 1$ convolution is applied firstly to the feature map to reduce the channel from 720 to 256. Then the feature map is split to separate feature tokens. For MVTec-AD \cite{bergmann2019mvtec} and CIFAR-10 \cite{krizhevsky2009learning}, there are 1024 and 64 feature tokens with the channel of 256, respectively. The position embedding is a learned embedding with the same size as the input feature tokens.

The transformer encoder follows the standard architecture in \cite{vaswani2017attention} with 4 layers. Each layer consists of a multi-head self-attention layer, a feed forward layer, and a shortcut connection with layer normalization. The head number in attention is 8. The architecture of the feed forward layer is shown as follows. 
\begin{center}
\begin{tabular}{cccccc}
\toprule
\textbf{Layer} & Input & FC1  & Relu  & FC2 \\
\midrule
\textbf{Output Size} & 256 & 1024  & 1024  & 256 \\
\midrule
\end{tabular} 
\end{center}
Besides, the position embedding is added in each self-attention layer rather than only in the first layer to keep more position information.

The transformer decoder also has 4 decoder layers. Each layer is composed of 2 parts, the self-attention part and the cross-attention part. The self-attention part includes a multi-head self-attention layer and a shortcut connection with layer normalization. The cross-attention part consists of a multi-head cross-attention layer, a feed forward layer, and a shortcut connection with layer normalization. The head number in both attention layers is set as 8. The architecture of the feed forward layer is the same as that in the transformer encoder. Also, the position embedding is added in all attention layers. The query embedding is a learned embedding with the same size as the input feature tokens. 

The outputs of the transformer have the same size as the inputs ($1024 \times 256$ for MVTec-AD, $64 \times 256$ for CIFAR-10). Then a $1 \times 1$ convolution is applied to increase the channel from 256 to 720. After reshape, we obtain the reconstructed feature map ($32 \times 32 \times 720$ for MVTec-AD, $8 \times 8 \times 720$ for CIFAR-10).

\textbf{Training configurations on MVTec-AD}.
In \textit{normal-sample-only case}, the backbone is frozen. The transformer is trained with $\mathcal{L}_{norm}$ in Eq. (3) for 500 epochs with batch size 16. AdamW optimizer \cite{AdamW} with weight decay $1 \times 10^{-4}$ is used. The learning rate is set as $1 \times 10^{-4}$ initially, and dropped by 0.1 after 400 epochs. In \textit{anomaly-available case}, the trained model in \textit{normal-sample-only case} is firstly loaded. The transformer is trained with $\mathcal{L}_{px}$ in Eq. (6) for 300 epochs. $\alpha$ in Eq. (6) is set as 0.003. The learning rate is initially set as $1 \times 10^{-4}$, and dropped by 0.1 after 200 epochs.

\textbf{Training configurations on CIFAR-10}.
In \textit{normal-sample-only case}, the details are the same as those in \textbf{MVTec-AD} except the image size and feature size described in \textbf{Backbone}. For more efficient training, the batch size is set as 128. In \textit{anomaly-available case}, the same implementations as \textbf{MVTec-AD} are adopted except the followings. Considering that the anomalies are image-level labeled in CIFAR-10 case, the transformer is trained with $\mathcal{L}_{img}$ in Eq. (8), where $\alpha$ and $k$ are selected as 0.003 and 20, respectively.

\section{More Visualization Results}\label{sec:results}

\begin{figure*}[!ht]
\begin{center}
   \includegraphics[width=1\linewidth]{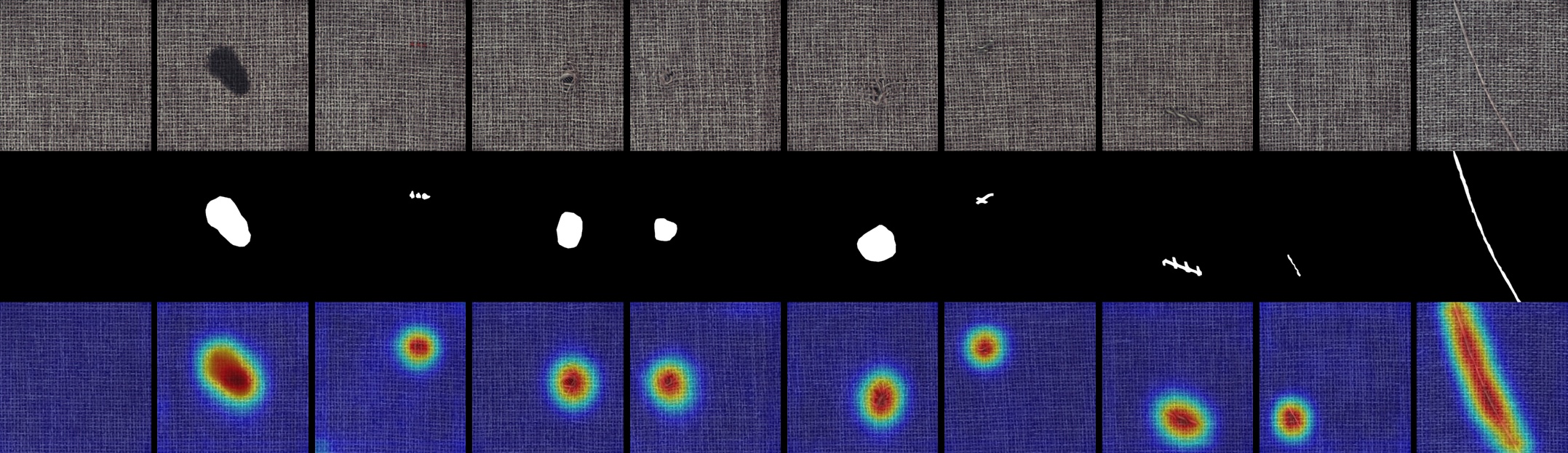}
\end{center}
\vspace{-12pt}
\caption{Anomaly detection results of carpet on MVTec-AD. From top to down: samples, ground-truth, and the anomaly score maps of ADTR. The first column is the normal sample.}
\label{fig:carpet}
\end{figure*}

\begin{figure*}[!ht]
\begin{center}
   \includegraphics[width=1\linewidth]{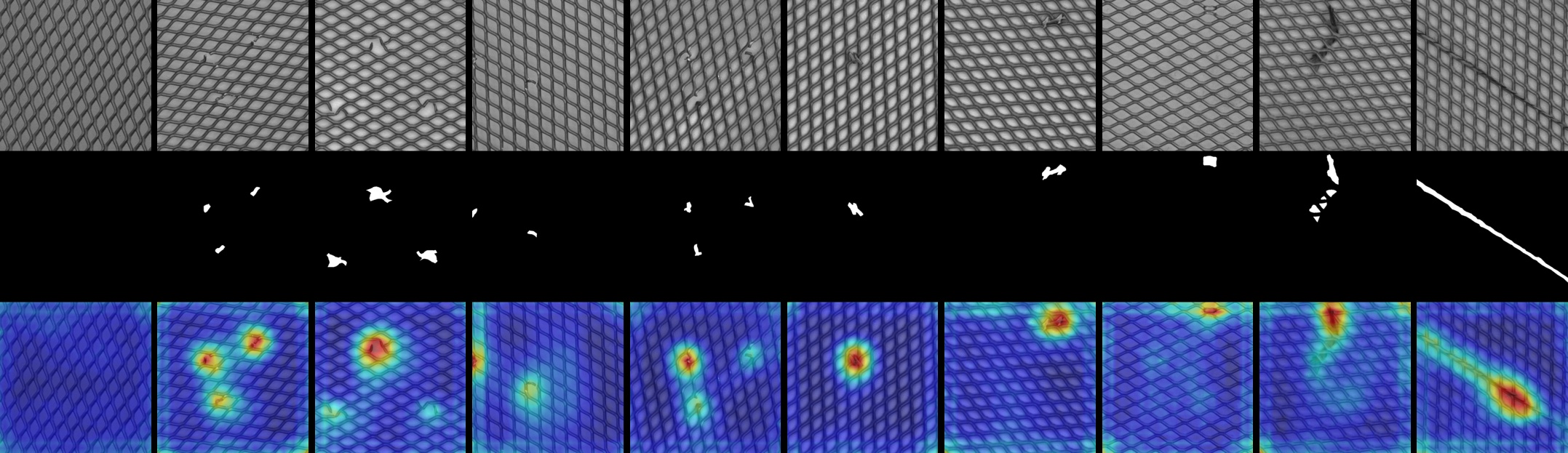}
\end{center}
\vspace{-11pt}
\caption{Anomaly detection results of grid on MVTec-AD. From top to down: samples, ground-truth, and the anomaly score maps of ADTR. The first column is the normal sample.}
\label{fig:grid}
\end{figure*}

\begin{figure*}[!ht]
\begin{center}
   \includegraphics[width=1\linewidth]{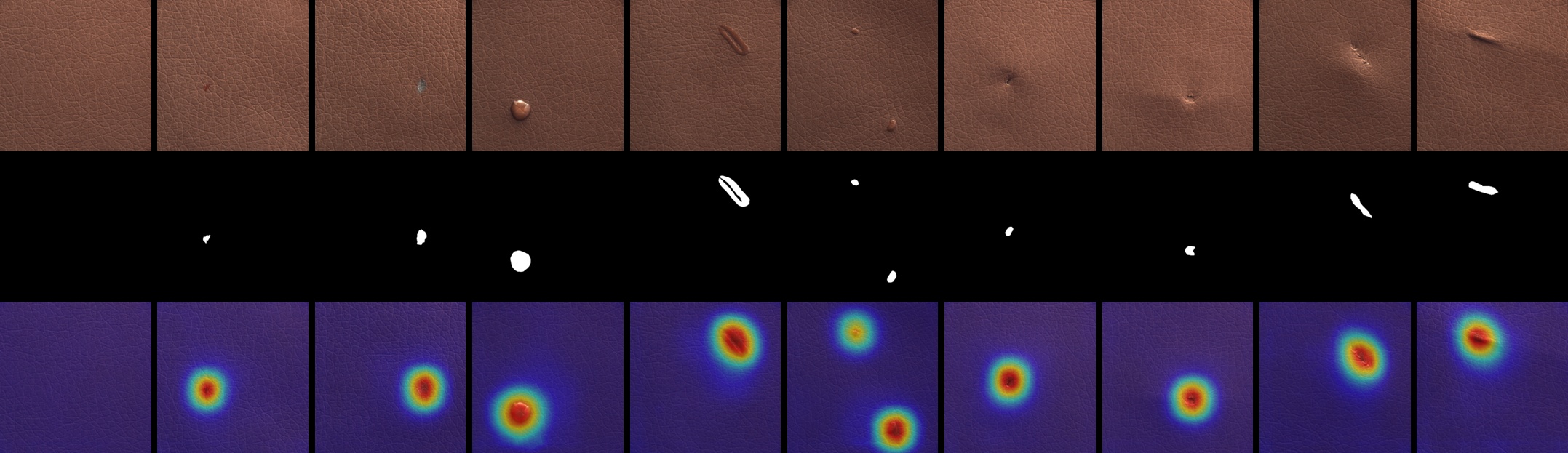}
\end{center}
\vspace{-11pt}
\caption{Anomaly detection results of leather on MVTec-AD. From top to down: samples, ground-truth, and the anomaly score maps of ADTR. The first column is the normal sample.}
\label{fig:leather}
\end{figure*}

\begin{figure*}[!ht]
\begin{center}
   \includegraphics[width=1\linewidth]{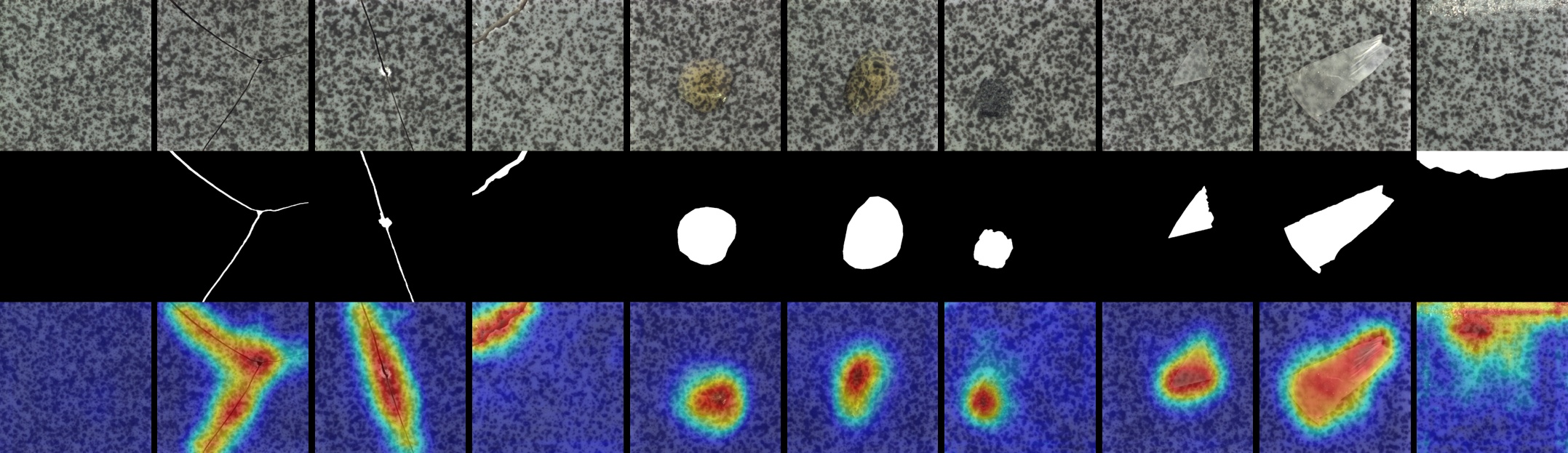}
\end{center}
\vspace{-10pt}
\caption{Anomaly detection results of tile on MVTec-AD. From top to down: samples, ground-truth, and the anomaly score maps of ADTR. The first column is the normal sample.}
\label{fig:tile}
\end{figure*}

\begin{figure*}[!ht]
\begin{center}
   \includegraphics[width=1\linewidth]{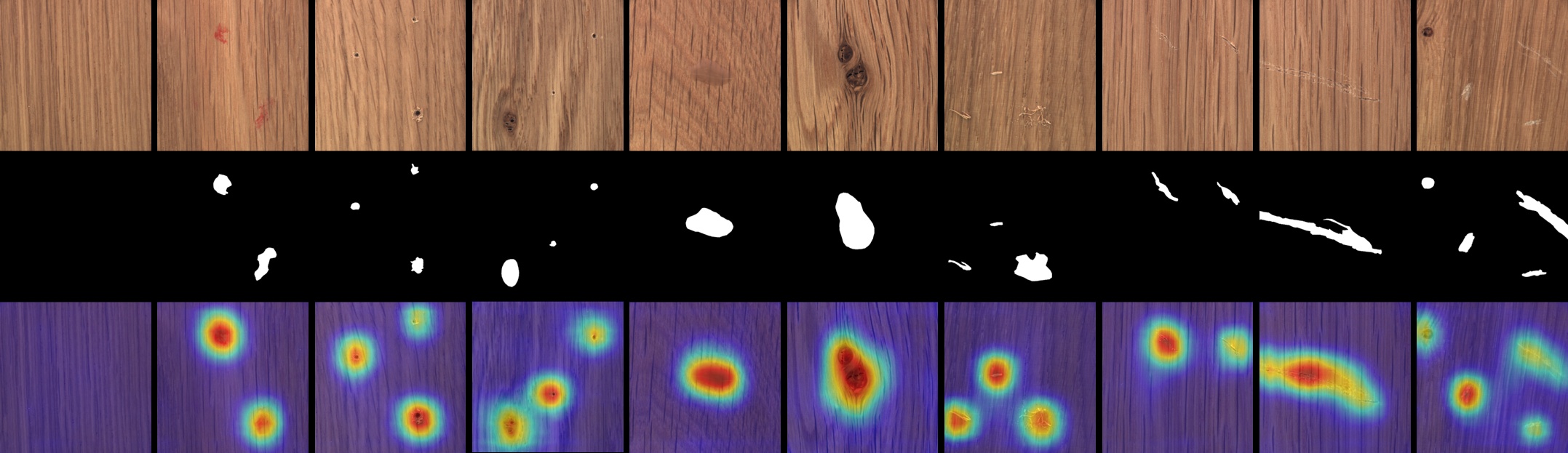}
\end{center}
\vspace{-10pt}
\caption{Anomaly detection results of wood on MVTec-AD. From top to down: samples, ground-truth, and the anomaly score maps of ADTR. The first column is the normal sample.}
\label{fig:wood}
\end{figure*}

\begin{figure*}[!ht]
\begin{center}
   \includegraphics[width=1\linewidth]{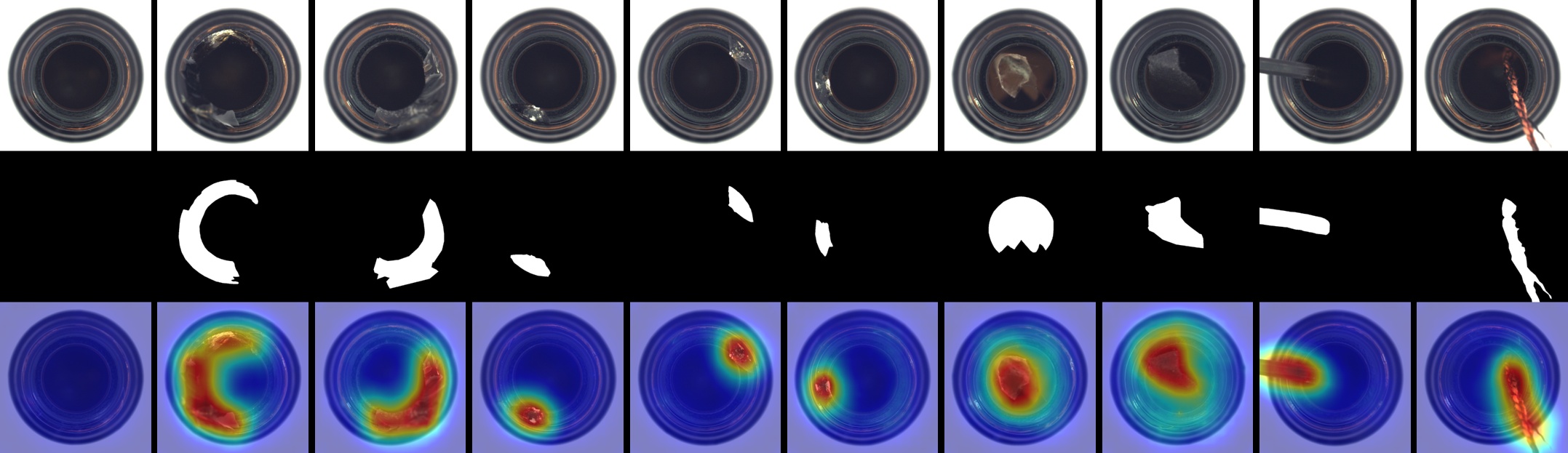}
\end{center}
\vspace{-10pt}
\caption{Anomaly detection results of bottle on MVTec-AD. From top to down: samples, ground-truth, and the anomaly score maps of ADTR. The first column is the normal sample.}
\label{fig:bottle}
\end{figure*}

\begin{figure*}[!ht]
\begin{center}
   \includegraphics[width=1\linewidth]{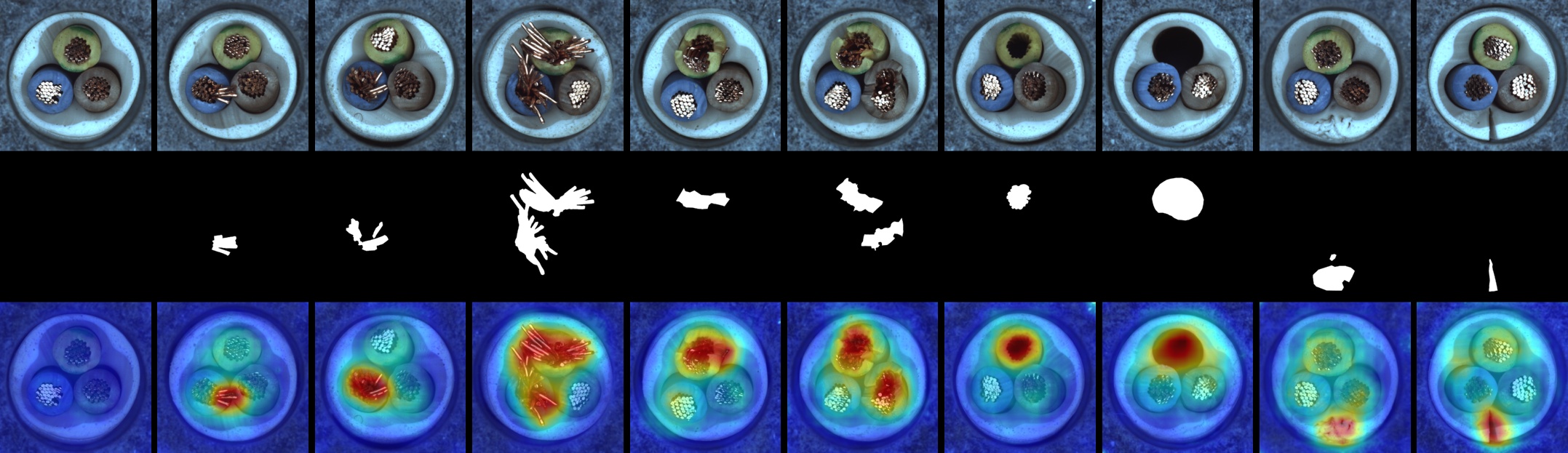}
\end{center}
\vspace{-10pt}
\caption{Anomaly detection results of cable on MVTec-AD. From top to down: samples, ground-truth, and the anomaly score maps of ADTR. The first column is the normal sample.}
\label{fig:cable}
\end{figure*}

\begin{figure*}[!ht]
\begin{center}
   \includegraphics[width=1\linewidth]{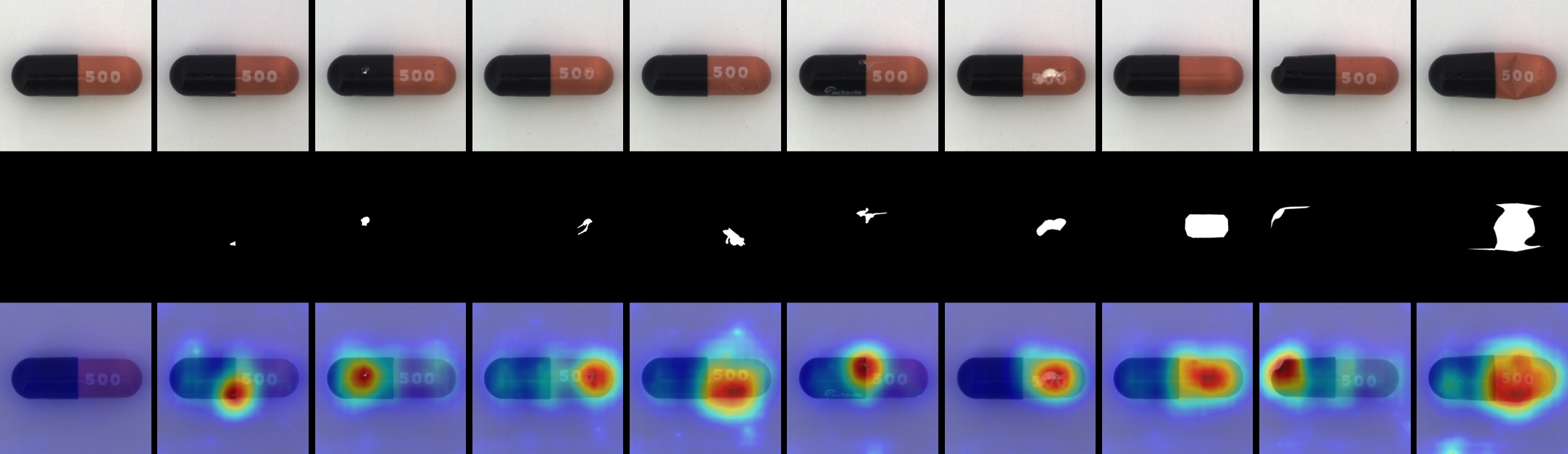}
\end{center}
\vspace{-10pt}
\caption{Anomaly detection results of capsule on MVTec-AD. From top to down: samples, ground-truth, and the anomaly score maps of ADTR. The first column is the normal sample.}
\label{fig:capsule}
\end{figure*}

\begin{figure*}[!ht]
\begin{center}
   \includegraphics[width=1\linewidth]{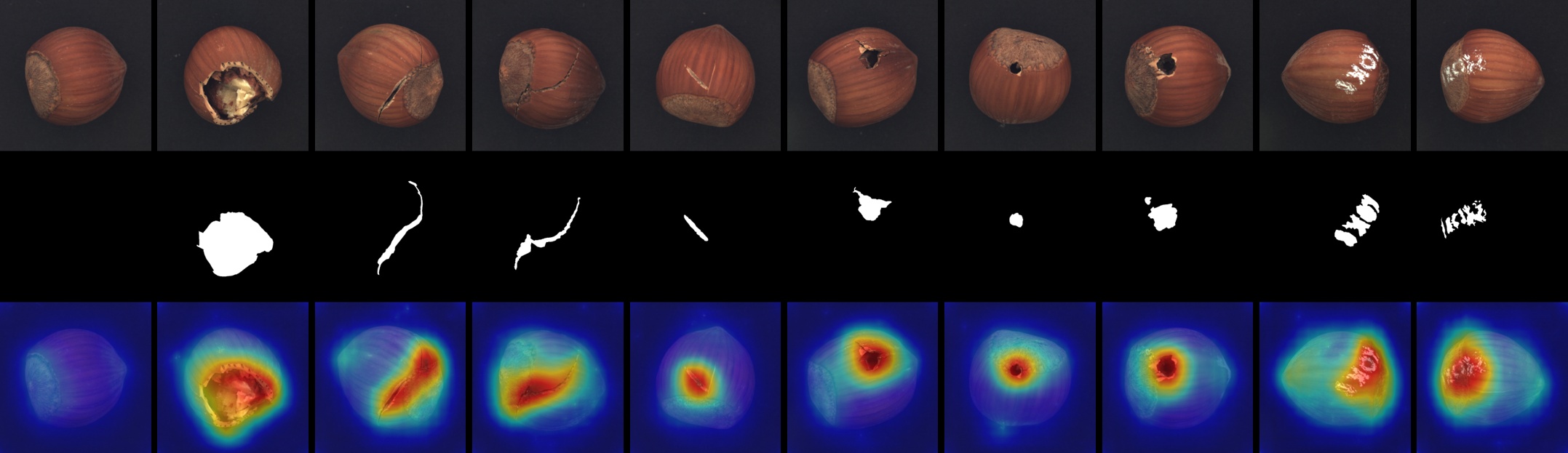}
\end{center}
\vspace{-10pt}
\caption{Anomaly detection results of hazelnut on MVTec-AD. From top to down: samples, ground-truth, and the anomaly score maps of ADTR. The first column is the normal sample.}
\label{fig:hazelnut}
\end{figure*}

\begin{figure*}[!ht]
\begin{center}
   \includegraphics[width=1\linewidth]{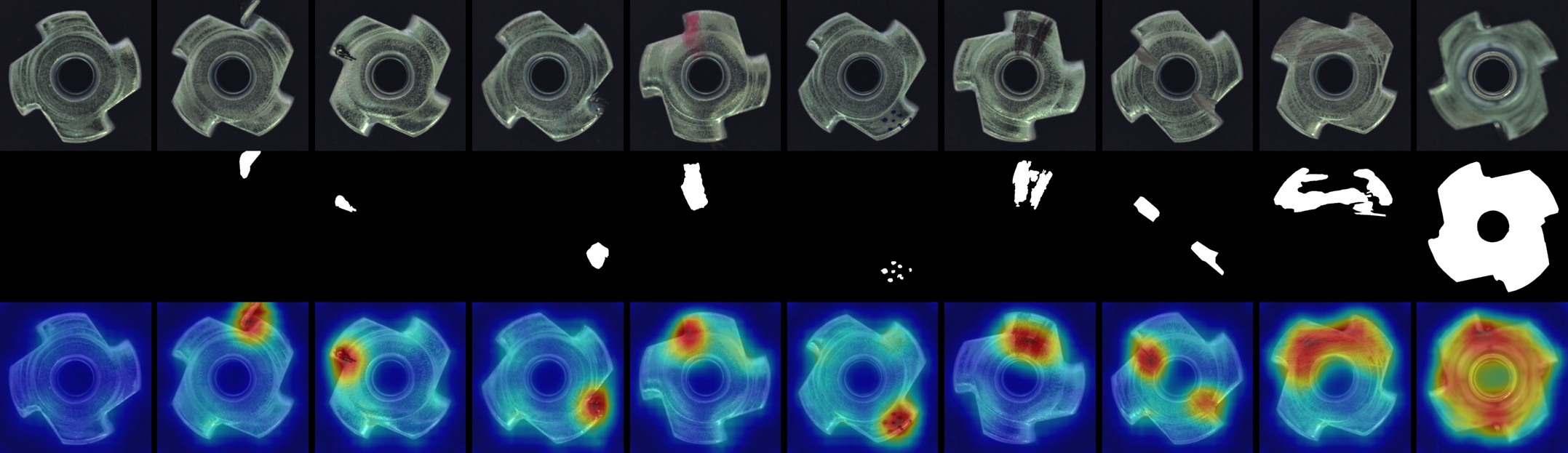}
\end{center}
\vspace{-10pt}
\caption{Anomaly detection results of metal nut on MVTec-AD. From top to down: samples, ground-truth, and the anomaly score maps of ADTR. The first column is the normal sample.}
\label{fig:metal_nut}
\end{figure*}

\begin{figure*}[!ht]
\begin{center}
   \includegraphics[width=1\linewidth]{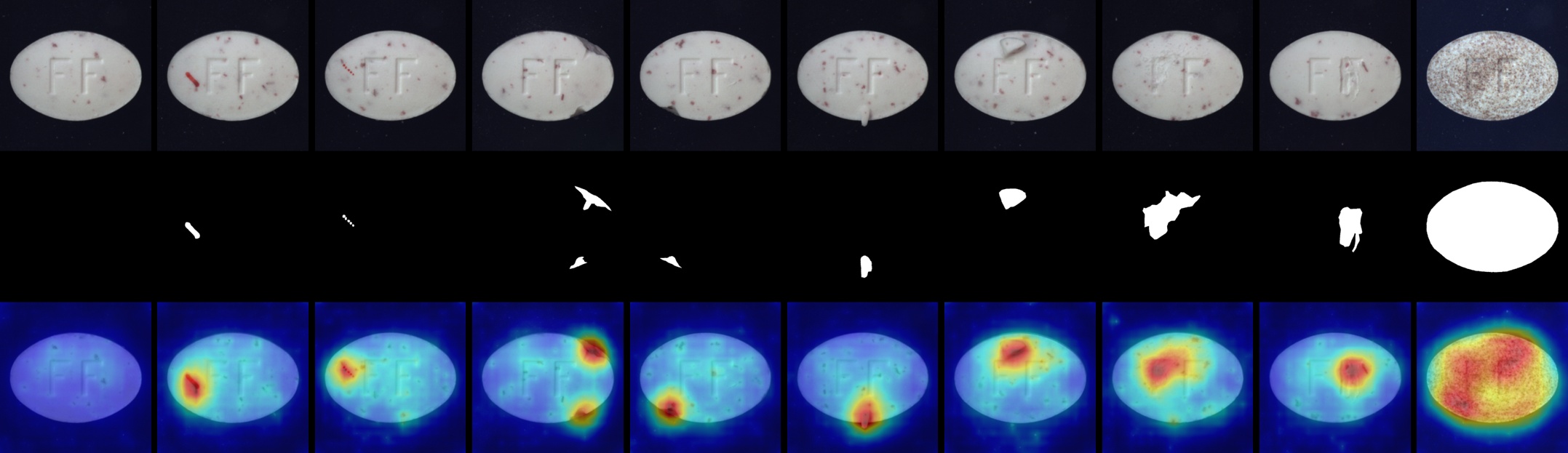}
\end{center}
\vspace{-10pt}
\caption{Anomaly detection results of pill on MVTec-AD. From top to down: samples, ground-truth, and the anomaly score maps of ADTR. The first column is the normal sample.}
\label{fig:pill}
\end{figure*}

\begin{figure*}[!ht]
\begin{center}
   \includegraphics[width=1\linewidth]{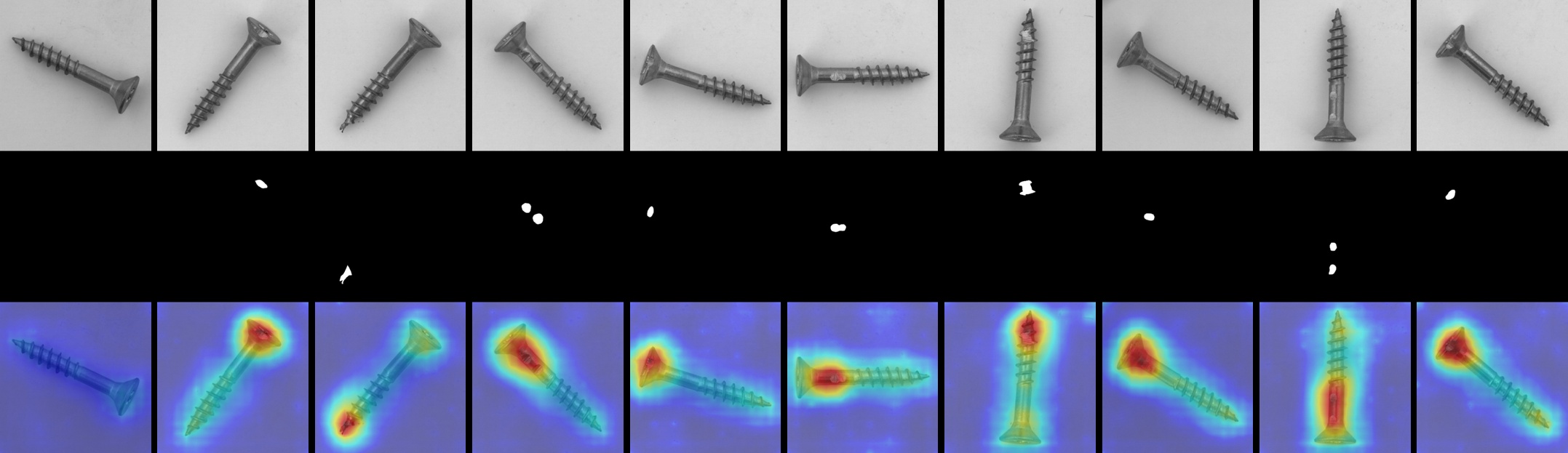}
\end{center}
\vspace{-10pt}
\caption{Anomaly detection results of screw on MVTec-AD. From top to down: samples, ground-truth, and the anomaly score maps of ADTR. The first column is the normal sample.}
\label{fig:screw}
\end{figure*}

\begin{figure*}[!ht]
\begin{center}
   \includegraphics[width=1\linewidth]{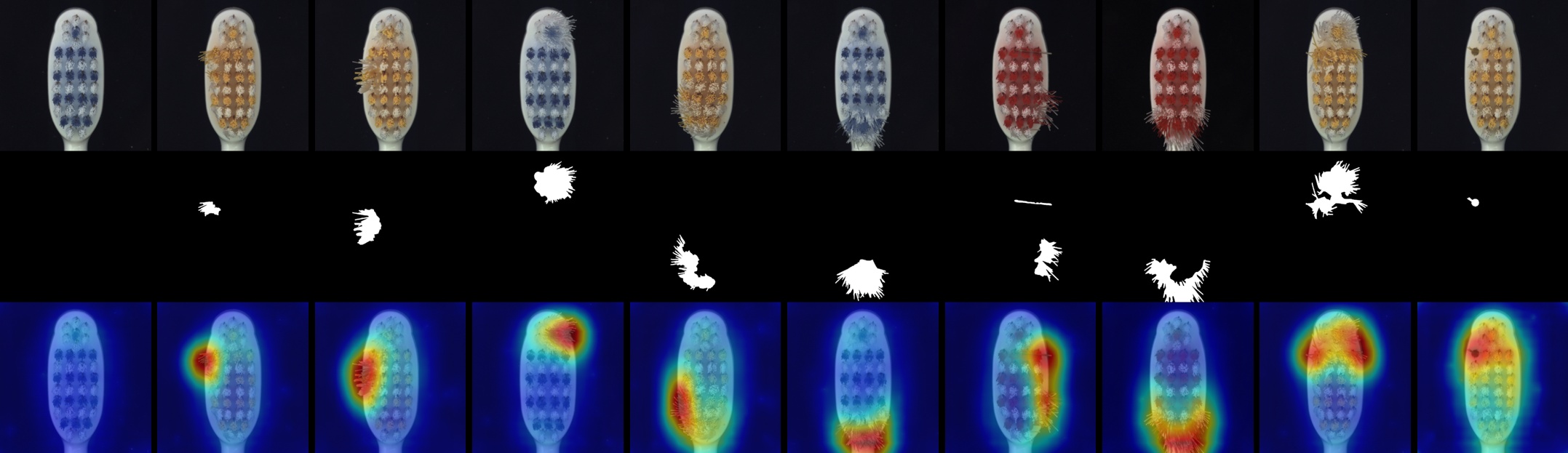}
\end{center}
\vspace{-10pt}
\caption{Anomaly detection results of toothbrush on MVTec-AD. From top to down: samples, ground-truth, and the anomaly score maps of ADTR. The first column is the normal sample.}
\label{fig:toothbrush}
\end{figure*}

\begin{figure*}[!ht]
\begin{center}
   \includegraphics[width=1\linewidth]{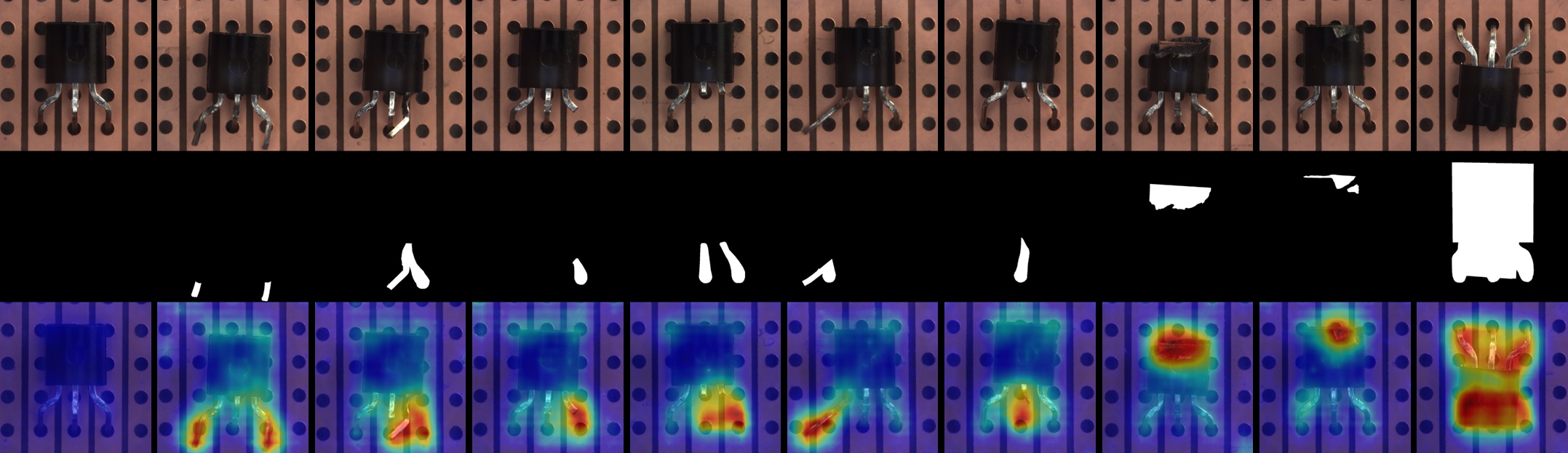}
\end{center}
\vspace{-10pt}
\caption{Anomaly detection results of transistor on MVTec-AD. From top to down: samples, ground-truth, and the anomaly score maps of ADTR. The first column is the normal sample.}
\label{fig:transistor}
\end{figure*}

\begin{figure*}[!ht]
\begin{center}
   \includegraphics[width=1\linewidth]{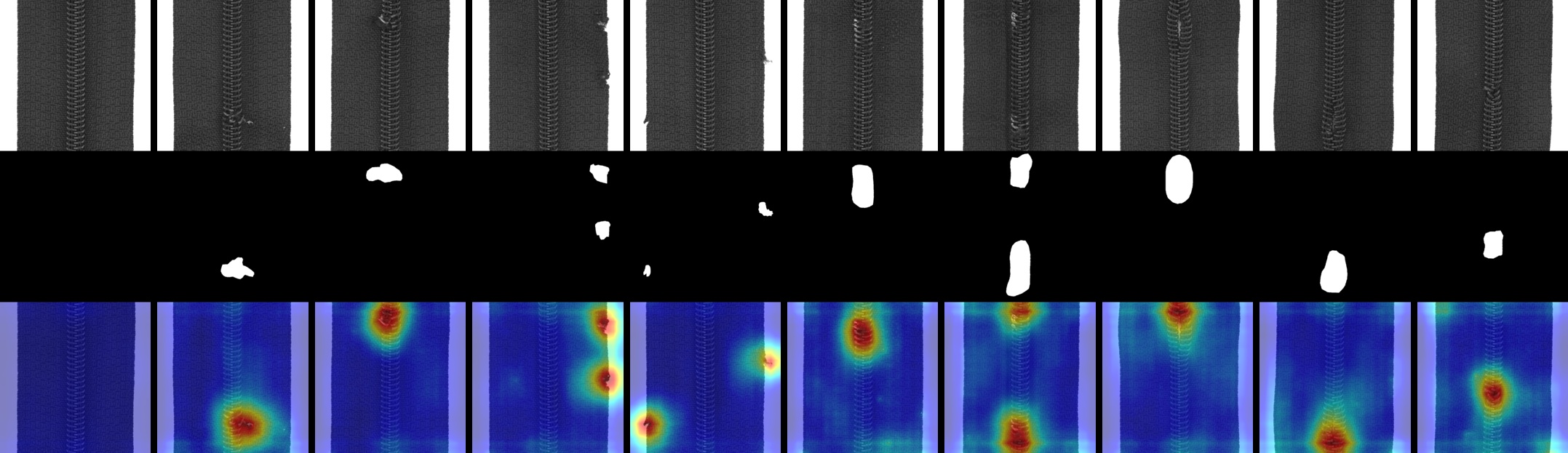}
\end{center}
\vspace{-10pt}
\caption{Anomaly detection results of zipper on MVTec-AD. From top to down: samples, ground-truth, and the anomaly score maps of ADTR. The first column is the normal sample.}
\label{fig:zipper}
\end{figure*}

\textbf{Qualitative results on MVTec-AD} are provided. 
These categories include: carpet (Fig. \ref{fig:carpet}), grid (Fig. \ref{fig:grid}), leather (Fig. \ref{fig:leather}), tile (Fig. \ref{fig:tile}), wood (Fig. \ref{fig:wood}), bottle
(Fig. \ref{fig:bottle}), cable (Fig. \ref{fig:cable}), capsule (Fig. \ref{fig:capsule}), hazelnut (Fig. \ref{fig:hazelnut}), metal nut (Fig. \ref{fig:metal_nut}), pill (Fig. \ref{fig:pill}), screw (Fig. \ref{fig:screw}), toothbrush (Fig. \ref{fig:toothbrush}), transistor (Fig. \ref{fig:transistor}), and zipper (Fig. \ref{fig:zipper}). Our approach could detect different kinds of anomalies in all categories with quite high localization accuracy. The performance of the proposed approach keeps stable in all these categories with various anomaly types, demonstrating strong generalization ability and robustness. Specifically, for both quite small anomalies (e.g. the second column in Fig. \ref{fig:capsule}) and quite large anomalies (e.g. the ninth column in Fig. \ref{fig:tile}), both single-kind anomalies (e.g. the second column in Fig. \ref{fig:leather}) and multi-kind combined anomalies (e.g. the last column in Fig. \ref{fig:wood}), both texture or color disorder (e.g. the second column in Fig. \ref{fig:carpet}) and misplacement (e.g. the last column in Fig. \ref{fig:transistor}), our approach could effectively detect all anomalies.

\begin{figure*}[!ht]
\begin{center}
   \includegraphics[width=1\linewidth]{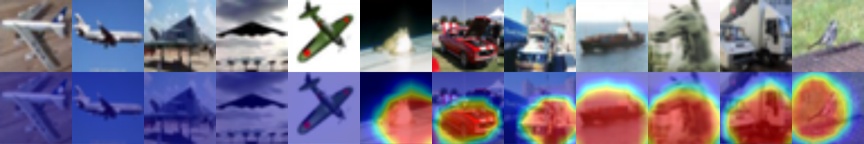}
\end{center}
\vspace{-15pt}
\caption{Anomaly detection results of airplane on CIFAR-10. From top to down: samples and the anomaly score maps of ADTR. Images from the first column to the fifth column are normal samples. }
\label{fig:airplane}
\end{figure*}

\begin{figure*}[!ht]
\begin{center}
   \includegraphics[width=1\linewidth]{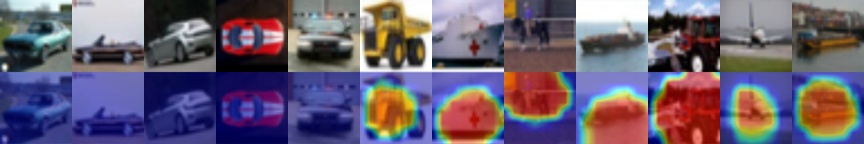}
\end{center}
\vspace{-15pt}
\caption{Anomaly detection results of automobile on CIFAR-10. From top to down: samples and the anomaly score maps of ADTR. Images from the first column to the fifth column are normal samples. }
\label{fig:auto}
\end{figure*}

\begin{figure*}[!ht]
\begin{center}
   \includegraphics[width=1\linewidth]{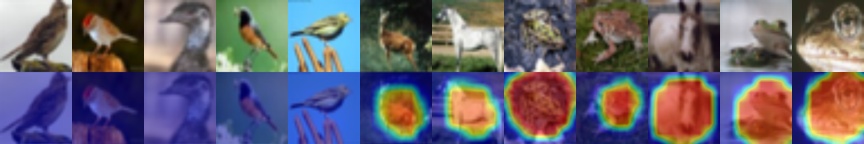}
\end{center}
\vspace{-15pt}
\caption{Anomaly detection results of bird on CIFAR-10. From top to down: samples and the anomaly score maps of ADTR. Images from the first column to the fifth column are normal samples. }
\label{fig:bird}
\end{figure*}

\begin{figure*}[!ht]
\begin{center}
   \includegraphics[width=1\linewidth]{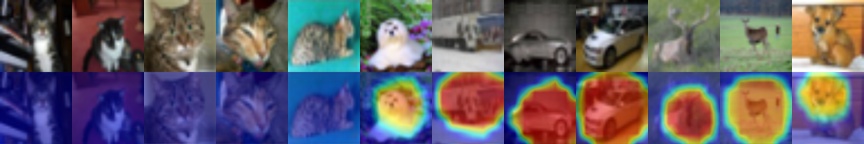}
\end{center}
\vspace{-15pt}
\caption{Anomaly detection results of cat on CIFAR-10. From top to down: samples and the anomaly score maps of ADTR. Images from the first column to the fifth column are normal samples. }
\label{fig:cat}
\end{figure*}

\begin{figure*}[!ht]
\begin{center}
   \includegraphics[width=1\linewidth]{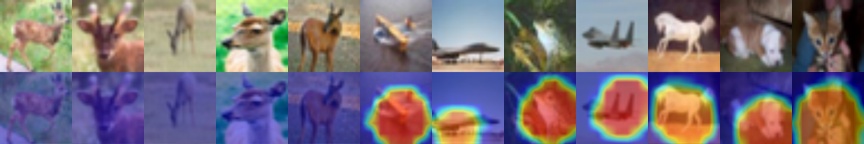}
\end{center}
\vspace{-15pt}
\caption{Anomaly detection results of deer on CIFAR-10. From top to down: samples and the anomaly score maps of ADTR. Images from the first column to the fifth column are normal samples. }
\label{fig:deer}
\end{figure*}

\begin{figure*}[!ht]
\begin{center}
   \includegraphics[width=1\linewidth]{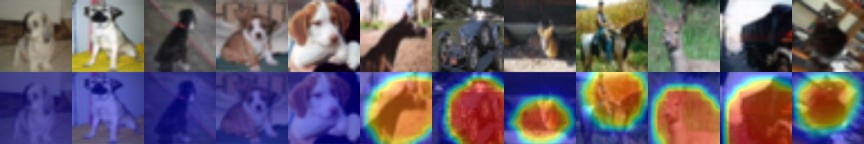}
\end{center}
\vspace{-15pt}
\caption{Anomaly detection results of dog on CIFAR-10. From top to down: samples and the anomaly score maps of ADTR. Images from the first column to the fifth column are normal samples. }
\label{fig:dog}
\end{figure*}

\begin{figure*}[!ht]
\begin{center}
   \includegraphics[width=1\linewidth]{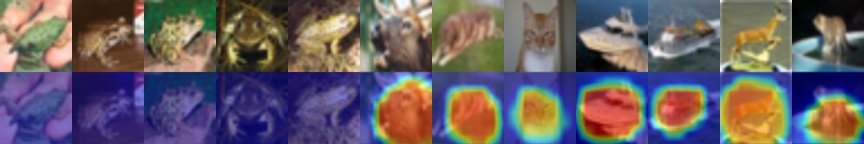}
\end{center}
\vspace{-15pt}
\caption{Anomaly detection results of frog on CIFAR-10. From top to down: samples and the anomaly score maps of ADTR. Images from the first column to the fifth column are normal samples. }
\label{fig:frog}
\end{figure*}

\begin{figure*}[!ht]
\begin{center}
   \includegraphics[width=1\linewidth]{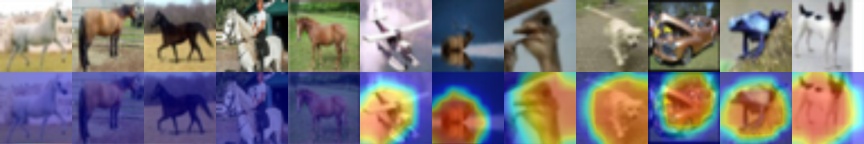}
\end{center}
\vspace{-15pt}
\caption{Anomaly detection results of horse on CIFAR-10. From top to down: samples and the anomaly score maps of ADTR. Images from the first column to the fifth column are normal samples. }
\label{fig:horse}
\end{figure*}

\begin{figure*}[!ht]
\begin{center}
   \includegraphics[width=1\linewidth]{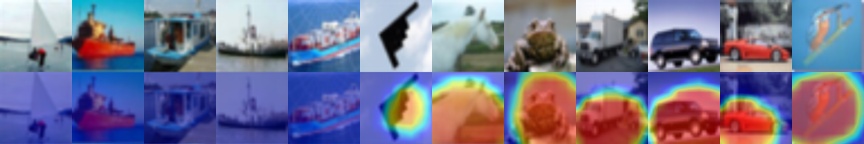}
\end{center}
\vspace{-15pt}
\caption{Anomaly detection results of ship on CIFAR-10. From top to down: samples and the anomaly score maps of ADTR. Images from the first column to the fifth column are normal samples. }
\label{fig:ship}
\end{figure*}

\begin{figure*}[!ht]
\begin{center}
   \includegraphics[width=1\linewidth]{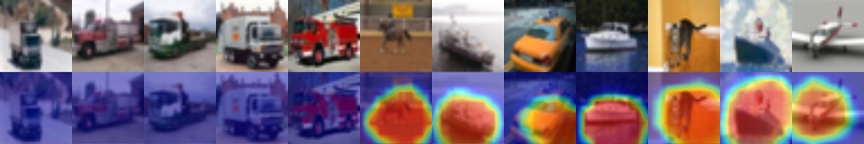}
\end{center}
\vspace{-15pt}
\caption{Anomaly detection results of truck on CIFAR-10. From top to down: samples and the anomaly score maps of ADTR. Images from the first column to the fifth column are normal samples. }
\label{fig:truck}
\end{figure*}

\textbf{Qualitative results on CIFAR-10} are given. 
These categories include: airplane (Fig. \ref{fig:airplane}), automobile (Fig. \ref{fig:auto}), bird (Fig. \ref{fig:bird}), cat (Fig. \ref{fig:cat}), deer (Fig. \ref{fig:deer}), dog
(Fig. \ref{fig:dog}), frog (Fig. \ref{fig:frog}), horse (Fig. \ref{fig:horse}), ship (Fig. \ref{fig:ship}), and truck (Fig. \ref{fig:truck}). Our approach could successfully detect various kinds of anomalies. Also, high anomaly scores mainly center on the anomaly objects rather than the backgrounds, which indicates that our approach detects anomalies based on the understanding of semantic features. In particular, even for anomalies that are very similar to normal samples, like the ``truck'' category when ``automobile'' category serves as normal samples (e.g. the sixth column in Fig. \ref{fig:auto}), the ``dog'' category when ``cat'' category serves as normal samples (e.g. the last column in Fig. \ref{fig:cat}), the ``horse'' category when ``deer'' category serves as normal samples (e.g. the tenth column in Fig. \ref{fig:deer}), our approach still successfully distinguishes these anomalies from normal samples.

\newpage

\end{document}